\documentclass[10pt,twocolumn,letterpaper]{article}

\usepackage{3dv}
\usepackage{times}
\usepackage{epsfig}
\usepackage{graphicx}
\usepackage{amsmath}
\usepackage{amssymb}
\usepackage{booktabs}
\usepackage{tabu} 
\usepackage{multirow} 
\usepackage[pagebackref=true,breaklinks=true,colorlinks,bookmarks=false]{hyperref}

\usepackage[capitalize]{cleveref}
\crefname{section}{Sec.}{Secs.}
\Crefname{section}{Section}{Sections}
\Crefname{table}{Table}{Tables}
\crefname{table}{Tab.}{Tabs.}

\threedvfinalcopy 

\ifthreedvfinal\fi

\begin{document}

\title{U-Attention to Textures: Hierarchical Hourglass Vision Transformer for Universal Texture Synthesis}

\author{Shouchang Guo\textsuperscript{1}\thanks{This work was mostly done when Shouchang Guo was an intern at Adobe Research.} , Valentin Deschaintre\textsuperscript{2}, Douglas Noll\textsuperscript{1}, Arthur Roullier\textsuperscript{2}
\\[0.1em]
\textsuperscript{1}University of Michigan, Ann Arbor, \textsuperscript{2}Adobe Research
}

\maketitle
\thispagestyle{empty}

\begin{abstract}
We present a novel U-Attention vision Transformer for universal texture synthesis. We exploit the natural long-range dependencies enabled by the attention mechanism to allow our approach to synthesize diverse textures while preserving their structures in a single inference. We propose a hierarchical hourglass backbone that attends to the global structure and performs patch mapping at varying scales in a coarse-to-fine-to-coarse stream. Completed by skip connection and convolution designs that propagate and fuse information at different scales, our hierarchical U-Attention architecture unifies attention to features from macro structures to micro details, and progressively refines synthesis results at successive stages. Our method achieves stronger 2$\times$ synthesis than previous work on both stochastic and structured textures while generalizing to unseen textures without fine-tuning. Ablation studies demonstrate the effectiveness of each component of our architecture.
\end{abstract}


\section{Introduction}
Texture synthesis aims at expanding a given texture image while preserving its structure and texture content. This finds application in multiple domains such as image inpainting or large-scale content generation.
In this work we leverage the attention mechanism~\cite{bahdanau2014neural} in Transformers~\cite{vaswani_attention_2017} to synthesize a 2$\times$ larger version of an input texture.
 
Recent learning-based approaches for texture synthesis showed impressive results using CNNs but require the training of a network for each texture \cite{Zhou2018NonstationaryTS,gatys2015texture}. These methods cannot generalize to unseen textures due to the locality of operations in CNNs~\cite{mardani2020}. To account for the global structural information required for the texture synthesis task, a different trend in recent work incorporates new designs for CNNs \cite{mardani2020,liu2020transposer} leveraging similarity maps or synthesizing textures in the Fourier domain. 
 
In this work, we focus on the emerging attention mechanism in Transformers that naturally exploits long-range structural dependencies required by texture synthesis and network generalization while reducing its computational overhead. Using our novel Transformer based architecture, our trained network is able to blend and match patches with various scales. This versatility lets our texture synthesis approach generalize to a broad range of texture patterns while remaining more computationally efficient than recent CNN-based networks~ \cite{mardani2020,liu2020transposer}.

We build a novel texture synthesis model that: (1) performs fast synthesis with a single forward inference, and (2) generalizes to different classes of textures with various amount of structure, with a single trained network. 

We propose a novel U-Attention net with multi-stage hierarchical hourglass Transformers as the generative model for texture synthesis. Our model presents two novel designs that distinguish it from other Transformer networks for image-to-image mapping, enabling fast texture synthesis for various types of textures. 
(1) Our hierarchical hourglass network exhibits an hourglass-like scale change of the input patches for different Transformer stages and enables progressive patch mapping with a coarse-fine-coarse scheme. (2) Each Transformer block is followed by a convolution and a down/up sampling layer, providing increased network capacity, with skip connection propagating pre-bottleneck information to later stages.

To summarize, our main contributions are:
\begin{itemize}
 
    \item A novel hierarchical hourglass backbone for coarse-to-fine and fine-back-to-coarse processing, allowing to apply self-attention at different scales and to exploit macro to micro structures. 
    
    \item Skip connections and convolutional layers between Transformer blocks, propagating and fusing high-frequency and low-frequency features from different Transformer stages.  

    \item A 2$\times$ texture synthesis method with a single trained network generalizing to various texture complexity in a single forward inference.
    
\end{itemize}

\section{Related Works}
\subsection{Algorithmic texture synthesis}
Texture synthesis aims at generating a larger extent of a given texture. This research area has been active for the past decades, as described in the survey of Raad et al.~\cite{surveyTextureSynthRaadDDM17}. More specifically, traditional methods follow two main approaches: statistics-based, using statistical descriptors of the original texture to generate the new image~\cite{Heeger1995PyramidbasedTA, Galerne2011, Galerne2012, Gilet2014LocalRN, Galerne2017TextonN, Heitz2018HighPerformanceBN}, and patch-based layout shuffling, leveraging available patches in the original texture to synthesize the new one~\cite{Efros1999TextureSB, quilting, Kaspar2015SelfTT}. The statistics-based methods can reproduce micro-structures well, but fail to preserve larger-scale patterns. Our method shares conceptual similarities with patch-based approaches but leverages multi-scale self-attention to evaluate patch relevance at a given position rather than a direct optimization.

\subsection{Deep-learning based texture synthesis}
More recently, deep learning based approaches have been proposed to synthesize textures. These methods can be divided into two families: iterative optimization approaches that use pretrained CNNs as feature extractors to match deep feature statistics of inputs and outputs~\cite{gatys2015texture, Isola2017ImagetoImageTW, WCT-NIPS-2017, Henzler20, rodriguez2019automatic}, and specialized networks which are trained per texture~\cite{Ulyanov2016TextureNF, Zhou2018NonstationaryTS, Shaham2019SinGANLA, Niklasson21, jetchev2016texture, Bergmann2017}. The latter produce impressive results but do not generalize to unseen textures without new training. Optimization methods using pretrained CNNs on the other hand often miss important texture elements, in particular in their structure~\cite{WCT-NIPS-2017, Henzler20, gatys2015texture} as they are based on convolutions and struggle to capture long-range dependencies in the textures. 

To alleviate these issues, Mardani et al.~\cite{mardani2020} proposed to synthesize textures in the Fourier domain in which global changes in image space are locally represented. Closer to our approach, Liu et al.~\cite{liu2020transposer} leverage a self-similarity map computed by comparing every possible combination of the shifted input texture at different positions. As opposed to previous work, we design a Transformer based architecture, leveraging the attention mechanism to naturally "attend" to different -- arbitrarily distant -- parts of textures at multiple scales. This new architecture allows our method to generalize to a large variety of textures and better handle structural elements.

\subsection{Transformers for images}

The Attention mechanism has been introduced in the context of natural language processing~\cite{bahdanau2014neural} as a way to solve the problem of long-range dependency on input sequences and was later adapted to the Transformer architecture~\cite{vaswani_attention_2017}. Motivated by the numerous successes in the field, it was soon adapted to computer vision tasks \cite{chen_generative_2020, parmar_image_2018}.

However, the attention mechanism is not straightforward to adapt to images, specifically due to its computational (quadratic) complexity. Consequently, recent research focused on improving the attention mechanism efficiency. Different approaches approximate the formula itself: the Performer \cite{choromanski_rethinking_2021} linearizes the softmax using the kernel trick, the Linformer \cite{wang_linformer_2020} lowers the rank of the attention matrix, the Perceiver \cite{Jaegle2021PerceiverGP} uses a recurrent update of a low dimensional latent representation, and the Synthesizer \cite{tay_synthesizer_2021} approximates the attention matrix itself. Another line of work relaxes the (spatial) range over which information is retrieved: Lambda Networks \cite{bello_lambdanetworks_2021} restrict the attention to a local neighborhood, Axial DeepLab \cite{wang_axial-deeplab_2020} restricts attention to specific directions, the Vision Transformer~\cite{dosovitskiy_image_2021} splits the image into crops instead of working at the pixel level.

Different visual tasks are tackled by recent works on image Transformers, including classification \cite{dosovitskiy_image_2021}, image processing \cite{chen2021pre}, super-resolution \cite{yang2020learning}, and video inpainting \cite{zeng2020learning}. Similar to these works we leverage the attention mechanism to image patches. For our universal texture synthesis goal, we propose a hierarchical hourglass architecture that 
progressively refines patch mapping with successive Transformer stages and unifies attention to multi-scales with an hourglass-like scale change between stages.

\section{U-Attention Model Architecture}
\begin{figure*}[t]
\centering
\includegraphics[width=0.96\textwidth]{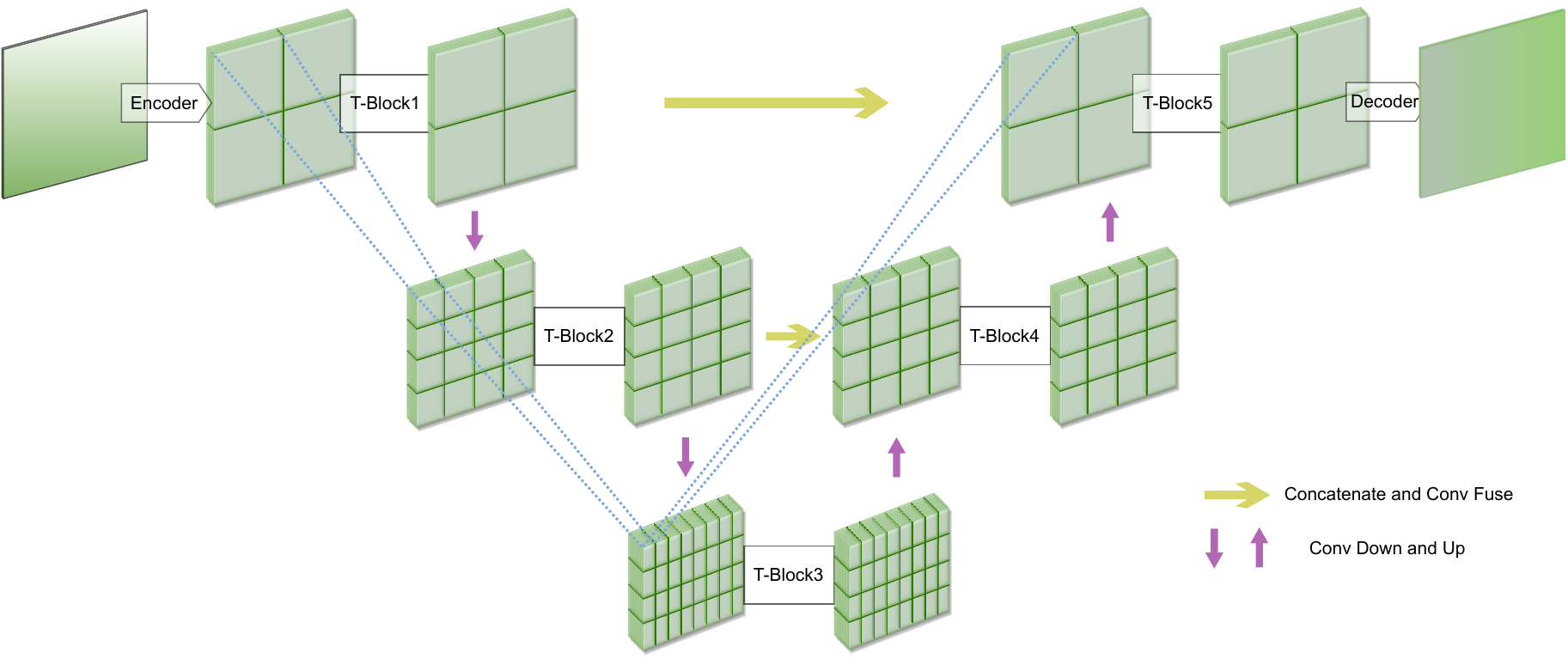}
\caption{
   Proposed U-Attention framework with hierarchical hourglass Transformers. 
   We introduce a multi-scale partition of the feature map between hierarchical Transformer blocks to form input patches of different scales for different Transformers.
   The input texture image is first projected into feature space by an encoder. We then leverage a succession of Transformer blocks, with up and down convolutions in between (purple arrows), processing the feature maps at different resolutions.
   Each Transformer block takes the whole feature maps as input, and we partition the feature maps to be sequences of patches of progressively smaller or larger sizes at consecutive stages of the network.
   Therefore, the input patch size of all the stages forms an hourglass-like scale change (dotted blue line), enabling attention to finer/coarser details at different attention steps. 
   Finally, we add skip connections that propagate and concatenate outputs from different previous stages as part of the inputs for later Transformer stages (yellow arrows).
   }\label{f1}
\vspace{-0.2cm}
\end{figure*}

\subsection{Overview}

\textbf{Problem statement}
Texture synthesis takes a small texture exemplar as input and generates a larger scale image of the same texture. Many texture synthesis approaches leverage patch matching and mapping at a given scale to enlarge texture content based on patches of the given exemplar texture ~\cite{Efros1999TextureSB,quilting,Kaspar2015SelfTT}. This leads to two challenges: (1) small scale patch mapping often fails to capture global structures and long-range dependencies in an input texture, and (2) choosing a scale for patch mapping requires manual tuning for different levels of structure depending on the input texture images. 

To take global structure into account, we propose to use attention mechanisms attending to the entire input image and able to process long-range dependencies. To automatically synthesize texture without manual scale parameter tuning, we propose a novel hierarchical attention network that unifies attention at different scales of texture details. 

\textbf{Architecture}
\cref{f1} presents our proposed U-Attention Network architecture. The input image to be outpainted is: (1) encoded by a set of convolutional layers into a feature domain before (2) being processed by our main hierarchical Transformer network and (3) being decoded back to the image domain by a set of convolutions.

Our main network consists of an odd number (e.g., 5) of Transformer blocks arranged in an hourglass shape for multi-scale and hierarchical patch mapping.
For each Transformer block $i$, we partition the current latent feature map of dimension ${H_i \times W_i \times C}$ into a sequence of $P_i^2$ patches of dimension $\frac{H_i}{P_i} \times \frac{W_i}{P_i} \times C$ as the input. 
The Transformer block processes the patch sequence and exploits long-range dependencies with a self-attention mechanism that attends to all the patches.
After processing, the transformed sequence of patches forms the transformed feature map, which is passed to the next stages of the model. 

The first 2 Transformer blocks are followed by strided convolutions that down-sample their inputs and enlarge the channel dimension, while outputs from the following 2 blocks are up-sampled and shrunk in the channel dimension.
We use skip connections, respectively between the outputs of Transformer blocks 1 and 2, and the inputs of Transformer blocks 4 and 5 (i.e., at the same resolution), allowing to propagate high frequency information extracted by the Transformer blocks before the bottleneck to be part of the input for later Transformer blocks.

We train our network using a structural $l_1$ loss, VGG-based perceptual and style losses \cite{liu2020transposer}, and a patch GAN loss as proposed in \cite{zeng2020learning,chang2019free}.


\subsection{Hierarchical hourglass vision Transformers}
\begin{figure}
\centering
\includegraphics[width=0.96\linewidth]{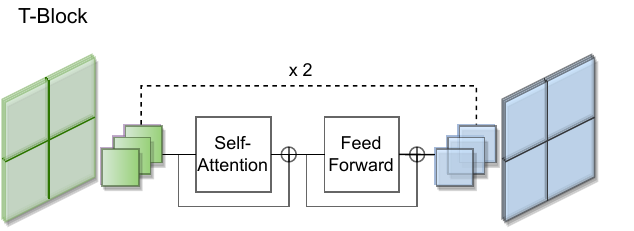}
\vspace{0.2cm}
\caption{
    Details of a single Transformer block (T-block). An input latent feature map is partitioned into a sequence of patches, and the input patch sequence is mapped to another patch sequence via a stack of 2 Transformer layers. The output sequence is then reshaped back into a whole feature map.
}\label{fig:ViTmodule}
\vspace{-0.2cm}
\end{figure}

\textbf{Transformer block}
Our Transformer block consists of 2 stacked Transformer layers, where each Transformer layer \cite{vaswani_attention_2017} contains attention, feed forward operation, residual connections, and normalization as illustrated in \cref{fig:ViTmodule}.

Each of our Transformer block $i$ operates on patches of latent space feature maps. The sequence of $P_i^2$ patches with dimensions $\frac{H_i}{P_i}\times\frac{W_i}{P_i}\times C$ is turned into a tensor 
$X \in \mathbb{R}^{P_i^2 \times \frac{H_i}{P_i} \times \frac{W_i}{P_i} \times C}$
and is processed via self-attention:
\begin{equation}
\textrm{self-attention}(X) = \sigma\left(\frac{W_Q(X)\, W_K(X)^\top}{\sqrt{d}} \right) W_V(X)
\label{eq:attn}
\end{equation}
where $\sigma$ is the softmax function and $\cdot^\top$ is the matrix transpose. $W_Q(\cdot)$, $W_K(\cdot)$, and $W_V(\cdot)$ denote operations that perform $1\times1$ convolutions along the channel dimension as in \cite{zhang2019self,zeng2020learning} to embed input patches to [queries, keys, and values], and further reshape the [queries, keys, and values] to matrices of dimension $P_i^2\,\times\,d$, where $d = \frac{H_i}{P_i}\,\frac{W_i}{P_i}\,C$.
The self-attention operation maps a sequence of patches to another sequence of patches with a learned attention map.
Specifically, the keys $W_K(X)$ can be thought of as \emph{content descriptors} broadcast by each input patch (e.g. ``I contain some brick texture"), the queries $W_Q(X)$ can be thought of as what each patch is \emph{interested in} (e.g. ``I contain half a brick texture and would need some brick texture"), and the values $W_V(X)$ are the actual quantities that are merged together to compute the output, based on the matching of keys and queries.  

Essentially, the Transformer block performs a learned patch mapping with global information, and the output sequences of patches are reshaped back to form new feature maps as inputs for the next stages.  

\textbf{Hierarchical network}
To unify attention at varying scales, we propose a multi-stage hierarchical design that progressively reduces/enlarges both the size of the feature maps in between consecutive Transformer blocks and the size of the patches attended to by the attention mechanisms. Indeed, the resolution of the attention mapping is determined by the spatial extent of the considered patches. If the patches are too small, the correlation calculation for attention can be noisy and computationally expensive. While if the patch is too large, the mapping would be too coarse.

With these considerations in mind, we design our network with an hourglass-shape scale change. Following each Transformer block -- except for the last one -- we add a 2$\times$ down or up convolution as shown in \cref{f1}.
These convolutions reduce (resp. enlarge) the input spatial dimension and increase (resp. decrease) the channel dimension.
Because the operations for attention are implemented with $1\times1$ convolutions along the channel dimensions, the progressively increasing (resp. decreasing) channel dimensions allow the attention to leverage deeper representations of the original image at different Transformer blocks of our model. 

\begin{figure}
\centering
\includegraphics[width=0.96\columnwidth]{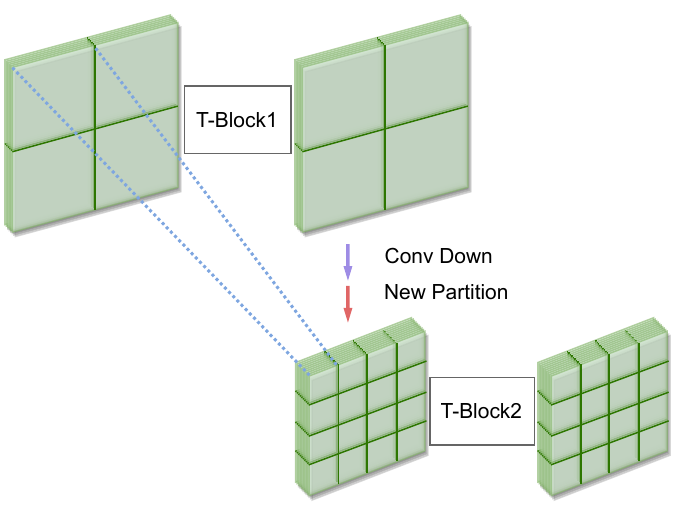}
\caption{
    Details of a transition between two Transformer stages that formulate the change of input patch scales for different Transformers. A feature map formed after a Transformer block (T-Block1) is downscaled by 2$\times$ with 4$\times$ enlarged channel dimension using a strided convolution, and then partitioned into 4$\times$ more patches as the input for the next Transformer block. Because the spatial extent is halved by the strided convolution, and the number of patches is doubled along each spatial dimension with the new partition, the input patches for the next Transformer block (T-Block2) are 4$\times$ smaller in each spatial dimension compared to the input patches for T-Block1.
    Each patch effectively represents 2$\times$ smaller regions than the previous step. 
}\label{fig:partition}
\vspace{-0.2cm}
\end{figure}

Importantly, as we reduce or augment the feature map resolution with convolutions, we further increase or decrease the spatial partitioning in each dimension by 2$\times$ to form longer or shorter patch sequences as inputs for different Transformer stages. Therefore, the size of the patches processed by the attention mechanism is divided or multiplied by 4$\times$ between consecutive Transformer blocks as shown in \cref{fig:partition}. 
This ensures that the Transformers can attend to different scales of features at different levels of the hourglass hierarchy. 
Specifically, as each spatial dimension is divided or multiplied by 2, while each patch dimension is divided or multiplied by 4, the network effectively attends to finer or coarser details with the footprint of attended patches being divided or multiplied by 2 at each stage.


We add skip connections to propagate high-frequency information from earlier stages to their mirrored later stage Transformer 
blocks as shown in yellow in \cref{f1}. Convolutional layers then fuse the skip connection information (yellow arrow) with the up-sampled output of the previous stage (purple arrow), forming the input of the next stage.

This U-Attention network unifies attention at different scales from different stages and builds progressive patch mapping
for both global pattern outline and local refinements of detailed structures.

\textbf{Coarse to fine to coarse}
We start from coarse scale attention to allow the first Transformer block to leverage the input texture and the overall structure. As the attention mechanism requires some information in the patches of the patch sequence to build a query and assess self-similarity, starting from small patches would require recursive processing that slowly builds the texture outward with little information on the global structure.
In our approach, at the coarsest level, the union of the patches covers the whole \--- signal free \--- area to synthesize and each patch contains part of the input texture. This enables our approach to synthesize a first rough scale texture with preserved original structure and relevant generation to be further refined in the following stages, as illustrated in \cref{fig:vis2}.

\textbf{Encoder-Decoder}
Before processing the inputs with our Transformer blocks, we encode them in feature space using a small 2 layer CNN, preserving the spatial extent but increasing the channel dimension. This lets the network manipulate a richer representation of images. Likewise, the last Transformer block outputs features that are then decoded by another 2 layer CNN similar to the encoder, returning features to the image space.

\subsection{Training and implementation details}
\label{sec:implementation}
The loss function $\mathcal{L}$ we use to train our model is a combination of structural $l_1$
loss in pixel space $\mathcal{L}_{l_1}$, perceptual loss $\mathcal{L}_{p}$ as the $l_1$ distance in the feature space of a pretrained VGG model, and a style loss $\mathcal{L}_s$ defined by the $l_1$ distance between the Gram matrices of the features from the same VGG model~\cite{gatys2015texture}. The perceptual loss and the style loss match the overall statistics of the texture without constraining a pixel perfect match. We also used a temporal-patch adversarial loss (T-PatchGAN) $\mathcal{L}_{GAN}$ \cite{chang2019free,zeng2020learning} that helps crafting realistic details. We set the batch dimension as the temporal dimension for the T-PatchGAN to distinguish spatial features of a batch as real or fake. 
Specifically, our complete loss is
\begin{equation}
\mathcal{L} = \lambda_{l_1} \mathcal{L}_{l_1} + \lambda_p \mathcal{L}_{p} + \lambda_s \mathcal{L}_{s} +  \lambda_{GAN} \mathcal{L}_{GAN}
\label{eq:loss}
\end{equation}
Similar to previous work~\cite{mardani2020,liu2020transposer}, we set $\lambda_{p} = 0.01$, $\lambda_{s} = 200$, and $\lambda_{GAN} = 0.1$.
We add a component $\mathcal{L}_{l_1}$ with $\lambda_{l_1} = 1$ to emphasize the structural synthesis and stabilize the network training.

We develop our model using Pytorch and train it using a single NVIDIA Tesla V100 GPU for 7 days. We use a batch size of 8 and train for 100 epochs with Adam \cite{kingma2015adam} optimizer at a constant learning rate of $0.001$. At inference time, our proposed network runs in 22 ms.

We use publicly available texture datasets with a wide variety of structures and patterns from 3 different sources \cite{cimpoi14describing,dai:synthesizability,kwitt_meerwald_2008}. There are about $27000$ texture images in total, and we randomly split them into $26400$ images for training, and about $600$ images for validation and testing. 
The images of different scales and resolutions are resized to $128\times128\times3$ as target images, and their $64\times64\times3$ central crops are zero-padded to $128\times128\times3$ as inputs.

\section{Experiments}
\begin{figure*}
\centering
\scalebox{0.68}{
\addtolength{\tabcolsep}{-4pt}
\begin{tabu}{cccccccc}
\multicolumn{1}{c}{} & \multicolumn{7}{c}{} \\
\rowfont{\Large} 
Input & Ours & WCT & pix2pixHD & Self Tuning & Texture CNN$^{*}$ & Naive Tiling  & Ground Truth \\ 
\includegraphics[width=0.09\textwidth]{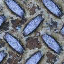} &
\includegraphics[width=0.18\textwidth]{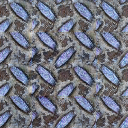} &
\includegraphics[width=0.18\textwidth]{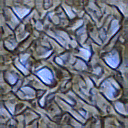} &
\includegraphics[width=0.18\textwidth]{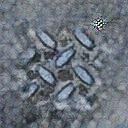} & 
\includegraphics[width=0.18\textwidth]{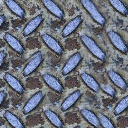} &
\includegraphics[width=0.18\textwidth]{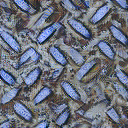} &
\includegraphics[width=0.18\textwidth]{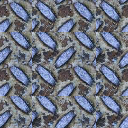} &
\includegraphics[width=0.18\textwidth]{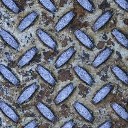} \\
\includegraphics[width=0.09\textwidth]{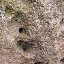} &
\includegraphics[width=0.18\textwidth]{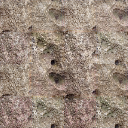} &
\includegraphics[width=0.18\textwidth]{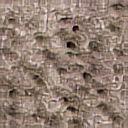} &
\includegraphics[width=0.18\textwidth]{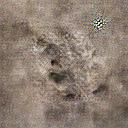} &
\includegraphics[width=0.18\textwidth]{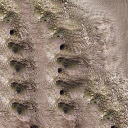} &
\includegraphics[width=0.18\textwidth]{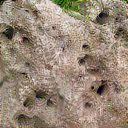} &
\includegraphics[width=0.18\textwidth]{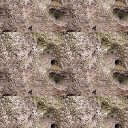} &
\includegraphics[width=0.18\textwidth]{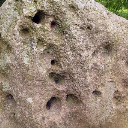} \\
\includegraphics[width=0.09\textwidth]{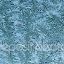} &
\includegraphics[width=0.18\textwidth]{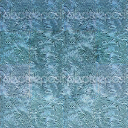} &
\includegraphics[width=0.18\textwidth]{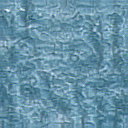} &
\includegraphics[width=0.18\textwidth]{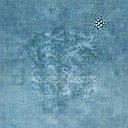} &
\includegraphics[width=0.18\textwidth]{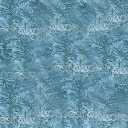} &
\includegraphics[width=0.18\textwidth]{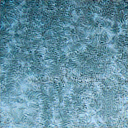} &
\includegraphics[width=0.18\textwidth]{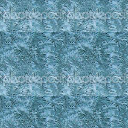} &
\includegraphics[width=0.18\textwidth]{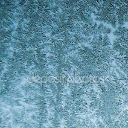} \\
\includegraphics[width=0.09\textwidth]{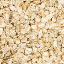} &
\includegraphics[width=0.18\textwidth]{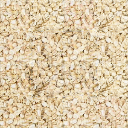} &
\includegraphics[width=0.18\textwidth]{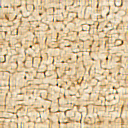} &
\includegraphics[width=0.18\textwidth]{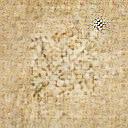} &
\includegraphics[width=0.18\textwidth]{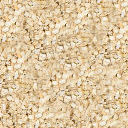} &
\includegraphics[width=0.18\textwidth]{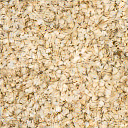} &
\includegraphics[width=0.18\textwidth]{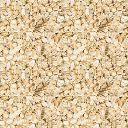} &
\includegraphics[width=0.18\textwidth]{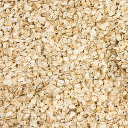} \\
\includegraphics[width=0.09\textwidth]{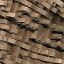} &
\includegraphics[width=0.18\textwidth]{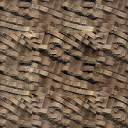} &
\includegraphics[width=0.18\textwidth]{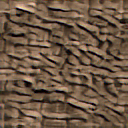} &
\includegraphics[width=0.18\textwidth]{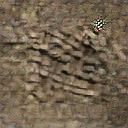} &
\includegraphics[width=0.18\textwidth]{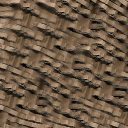} &
\includegraphics[width=0.18\textwidth]{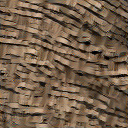} &
\includegraphics[width=0.18\textwidth]{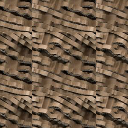} &
\includegraphics[width=0.18\textwidth]{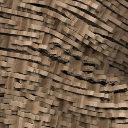} \\
\includegraphics[width=0.09\textwidth]{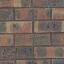} &
\includegraphics[width=0.18\textwidth]{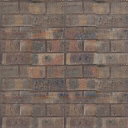} &
\includegraphics[width=0.18\textwidth]{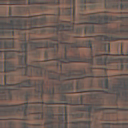} &
\includegraphics[width=0.18\textwidth]{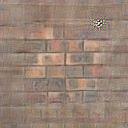} &
\includegraphics[width=0.18\textwidth]{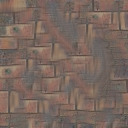} &
\includegraphics[width=0.18\textwidth]{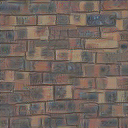} &
\includegraphics[width=0.18\textwidth]{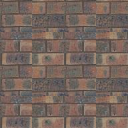} &
\includegraphics[width=0.18\textwidth]{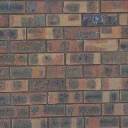} 
\end{tabu}
}
\vspace{0.2cm}
\caption{Comparison to previous works. We can see that our method does not suffer from structure randomness as WCT~\cite{WCT-NIPS-2017} or Texture CNN~\cite{gatys2015texture}. Our approach is also more robust in avoiding repetition artifacts as seen in Self Tuning~\cite{Kaspar2015SelfTT} or blurry boundaries like pix2pixHD~\cite{wang2018pix2pixHD}. We also show that our method suffers less from the strong discontinuities introduced by naive tiling.}
\label{fig:results}
\vspace{-0.1cm}
\end{figure*}

\begin{table}
    \centering
    \scalebox{0.95}{
    \begin{tabular}{|l|cccc|}
    \hline
    & {SSIM} & LPIPS & c-LPIPS & \begin{tabular}[c]{@{}c@{}}Inference \\[-0.3em] speed (s)\end{tabular}\\
    \hline\hline
    Naive Tiling & 0.104 & 0.529 & 0.500 & - \\
    WCT & 0.179 & 0.401 & 0.350 & 0.7\\ 
    pix2pixHD & 0.132 & 0.526 & 0.497 & 0.011\\ 
    Self Tuning & 0.104 & 0.540 & 0.514 & 150\\
    Texture CNN$^{*}$ & 0.199 & \textbf{0.272} & 0.256 & 100 \\
    \textbf{Ours} & \textbf{0.333} & \textbf{0.272} & \textbf{0.152} & 0.022 \\
    \hline
    \end{tabular}
    }
    \vspace{0.3cm}
    \caption{Quantitative comparison to previous work. Our approach performs better on structural and perceptual metrics. $^{*}$TextureCNN aims at the generation of a new texture variation from the entire ground truth image.}
    \label{tab:eval}
    \vspace{-0.1cm}
\end{table}


\begin{figure*}
\centering
\scalebox{0.68}{
\addtolength{\tabcolsep}{-4pt}
\begin{tabu}{cccccc}
\multicolumn{1}{c}{} & \multicolumn{5}{c}{} \\
\rowfont{\Large} Input & Baseline & Pyramid & Simplified Hourglass & No GAN Loss & Ours full \\ 
\includegraphics[width=0.12\textwidth]{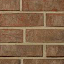} &
\includegraphics[width=0.24\textwidth]{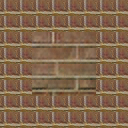} &
\includegraphics[width=0.24\textwidth]{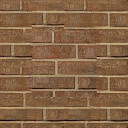} &
\includegraphics[width=0.24\textwidth]{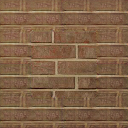} & 
\includegraphics[width=0.24\textwidth]{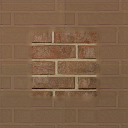} &
\includegraphics[width=0.24\textwidth]{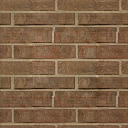} \\
\includegraphics[width=0.12\textwidth]{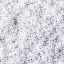} &
\includegraphics[width=0.24\textwidth]{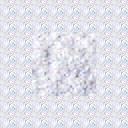} &
\includegraphics[width=0.24\textwidth]{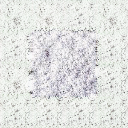} &
\includegraphics[width=0.24\textwidth]{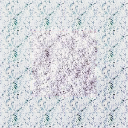} & 
\includegraphics[width=0.24\textwidth]{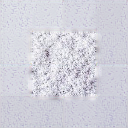} &
\includegraphics[width=0.24\textwidth]{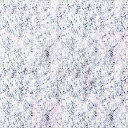} \\
\includegraphics[width=0.12\textwidth]{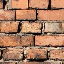} &
\includegraphics[width=0.24\textwidth]{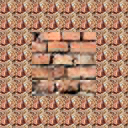} &
\includegraphics[width=0.24\textwidth]{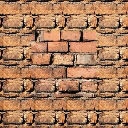} &
\includegraphics[width=0.24\textwidth]{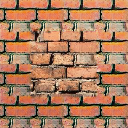} & 
\includegraphics[width=0.24\textwidth]{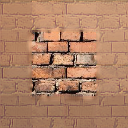} &
\includegraphics[width=0.24\textwidth]{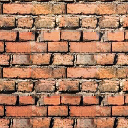} \\
\includegraphics[width=0.12\textwidth]{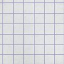} &
\includegraphics[width=0.24\textwidth]{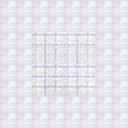} &
\includegraphics[width=0.24\textwidth]{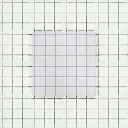} &
\includegraphics[width=0.24\textwidth]{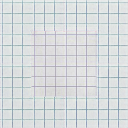} & 
\includegraphics[width=0.24\textwidth]{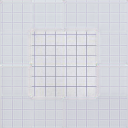} &
\includegraphics[width=0.24\textwidth]{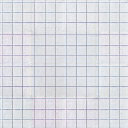} \\
\includegraphics[width=0.12\textwidth]{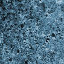} &
\includegraphics[width=0.24\textwidth]{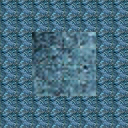} &
\includegraphics[width=0.24\textwidth]{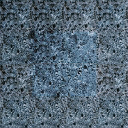} &
\includegraphics[width=0.24\textwidth]{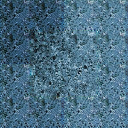} & 
\includegraphics[width=0.24\textwidth]{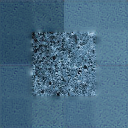} &
\includegraphics[width=0.24\textwidth]{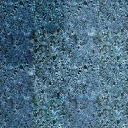} \\
\includegraphics[width=0.12\textwidth]{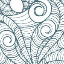} &
\includegraphics[width=0.24\textwidth]{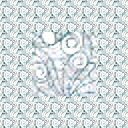} &
\includegraphics[width=0.24\textwidth]{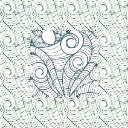} &
\includegraphics[width=0.24\textwidth]{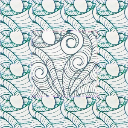} & 
\includegraphics[width=0.24\textwidth]{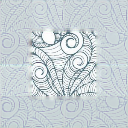} &
\includegraphics[width=0.24\textwidth]{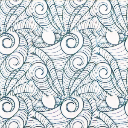} \\
\end{tabu}
}
\vspace{0.2cm}
\caption{Qualitative results for our ablation study. We can see that the baseline cannot attend to multiple attention scales. The Pyramid approach allows to attend to multiple scales, but cannot properly fuse the multi-scale attention information. The simplified hourglass does not have convolutions between Transformer blocks, failing to handle the overall structure and color coherence. Finally we ablate our GAN loss and show that it significantly improves the perceptual quality of the results.}
\label{fig:ablation}
\vspace{-0.1cm}
\end{figure*}

\begin{table}
    \centering
    \scalebox{0.95}{
    \begin{tabular}{|l|ccc|}
    \hline
    & {SSIM} & LPIPS & c-LPIPS \\
    \hline\hline
    Baseline & 0.256 & 0.388 & 0.294 \\
    Pyramid & 0.325 & 0.276 & 0.154 \\
    Hourglass & 0.334 & 0.284 & 0.156 \\
    No GAN loss & \textbf{0.381} & 0.350 & 0.204 \\
    \textbf{Ours} & 0.333 & \textbf{0.272} & \textbf{0.152} \\
    \hline
    \end{tabular}
    }
    \vspace{0.3cm}
    \caption{Qualitative evaluation of the ablation study. We see that our full approach better matches the perceptual appearance of the target (LPIPS/c-LPIPS). While the version without adversarial loss performs a little better when evaluated with SSIM, its perceptual appearance is significantly worse, as particularly visible in the qualitative experiment in \cref{fig:ablation}. With respect to structure, the pyramid cascaded Transformer seem to provide the strongest benefit.}
    \label{tab:ablation}
    \vspace{-0.1cm}
\end{table}


\begin{figure}
\def\arraystretch{0.5}
\addtolength{\tabcolsep}{-5pt}
\centering
\begin{tabu}{ccc}
\multicolumn{1}{c}{Input} 
& \multicolumn{1}{c}{Ours} 
& Ground Truth \\ 
\multicolumn{1}{c}{\includegraphics[height=0.06\textheight]{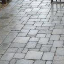}} 
& \multicolumn{1}{c}{\includegraphics[height=0.12\textheight]{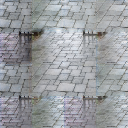}}
& {\includegraphics[height=0.12\textheight]{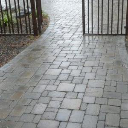}} \\
\multicolumn{1}{c}{\includegraphics[height=0.06\textheight]{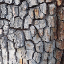}} 
& \multicolumn{1}{c}{\includegraphics[height=0.12\textheight]{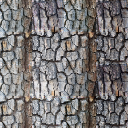}}
& {\includegraphics[height=0.12\textheight]{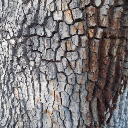}} \\ 
\multicolumn{1}{c}{\includegraphics[height=0.06\textheight]{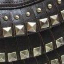}} 
& \multicolumn{1}{c}{\includegraphics[height=0.12\textheight]{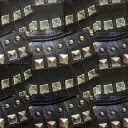}}
& {\includegraphics[height=0.12\textheight]{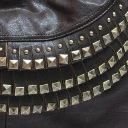}} \\ 
\multicolumn{1}{c}{\includegraphics[height=0.06\textheight]{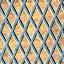}} 
& \multicolumn{1}{c}{\includegraphics[height=0.12\textheight]{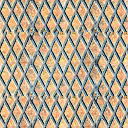}}
& {\includegraphics[height=0.12\textheight]{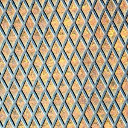}} \\ \multicolumn{1}{c}{\includegraphics[height=0.06\textheight]{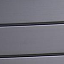}} 
& \multicolumn{1}{c}{\includegraphics[height=0.12\textheight]{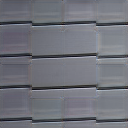}}
& {\includegraphics[height=0.12\textheight]{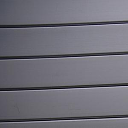}} \\
\end{tabu}
\vspace{0.2cm}
\caption{Example for limitations of our method. As we rely on patch mapping, our method does not handle well perspective or non periodic inputs  (rows 1-3). With the limitation of patch operations, in some cases the alignment cannot be well corrected by our current approach (rows 4-5).}
\label{fig:lim}
\vspace{-0.2cm}
\end{figure}

\subsection{Evaluation metrics}

For evaluation, we randomly select 400 images from the testing set and evaluate the synthesis performance based on qualitative results and quantitative metrics including structural similarity index measure (SSIM) \cite{wang2004image} and learned perceptual image patch similarity (LPIPS) \cite{zhang2018perceptual}. 
SSIM and LPIPS measure structural and perceptual similarity between image pairs and we use c-LPIPS which operates on crops of images, for texture synthesis evaluation. We compute c-LPIPS on 8 random crops of the target and output images.

\subsection{Comparisons and results}

We compare our method to the baseline and state-of-the-art, publicly available, approaches: 
(1) naive tiling, which tiles the input to the relative locations; 
(2) WCT \cite{WCT-NIPS-2017}, a style transfer method  using the small texture input as the style image and a 2$\times$ larger random noise image as the content image;
(3) pix2pixHD \cite{wang2018pix2pixHD}, the state-of-the-art image-to-image translation GAN network, which we trained with the same input and output pairs as for our work; 
(4) self tuning \cite{Kaspar2015SelfTT}, the state-of-the-art optimization method; 
(5) texture CNN$^*$ \cite{gatys2015texture} which uses the ground-truth image as input style and a noise image of the same size as the content image.

As presented in \cref{fig:results}, our model outperforms other approaches with generalization to unseen textures, and better preserves global structures including global color shift as shown in the third row and brick wall structures in the last row.
We present the numerical evaluation in \cref{tab:eval} and demonstrate that our method outperforms previous work on structural (SSIM, higher is better) and perceptual (LPIPS and c-LPIPS, lower is better) metrics. 

\subsection{Attention visualizations}

\begin{figure}
\def\arraystretch{0.5}
\addtolength{\tabcolsep}{5pt}
\centering
\begin{tabu}{ccc}
\multicolumn{1}{c}{Attention Map} & \multicolumn{1}{c}{Input} & Output \\ 
\multicolumn{3}{c}{\includegraphics[height=0.115\textheight]{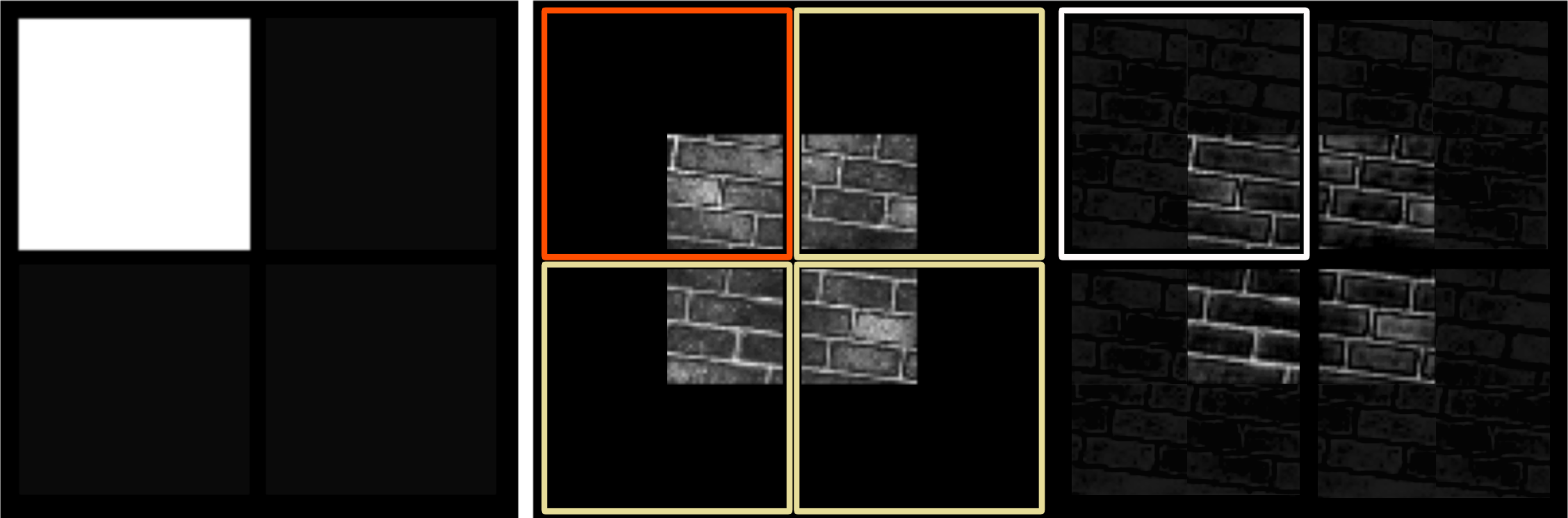}} \\ 
\multicolumn{3}{c}{\includegraphics[height=0.115\textheight]{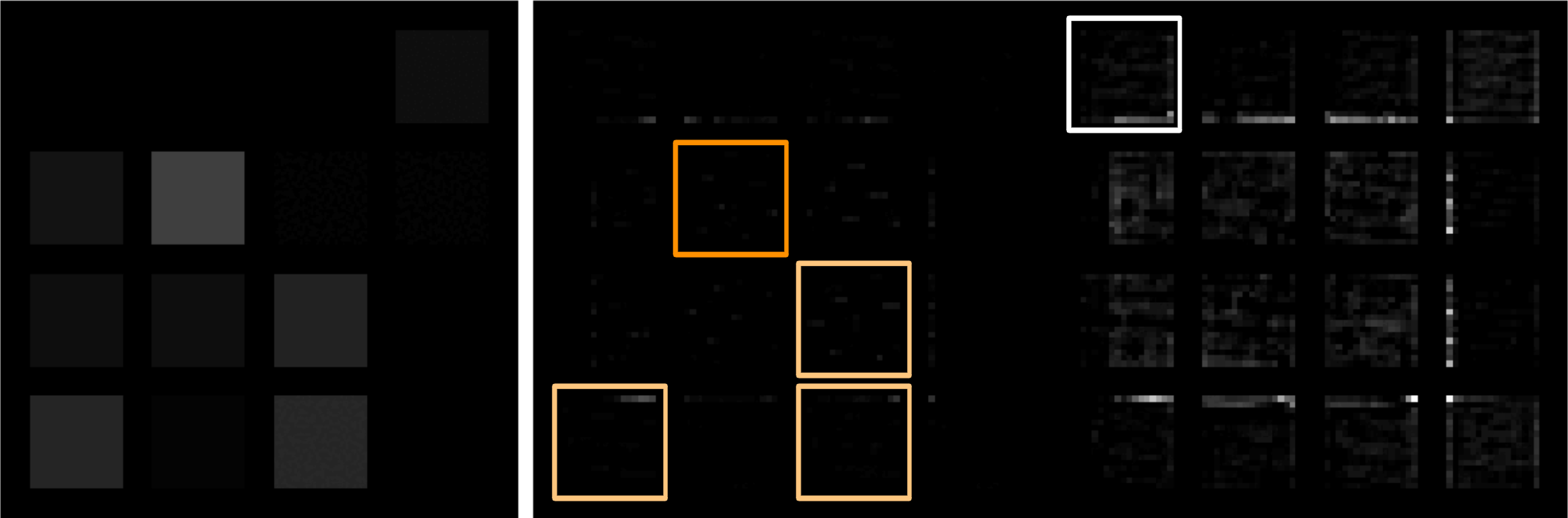}} \\ 
\multicolumn{3}{c}{\includegraphics[height=0.115\textheight]{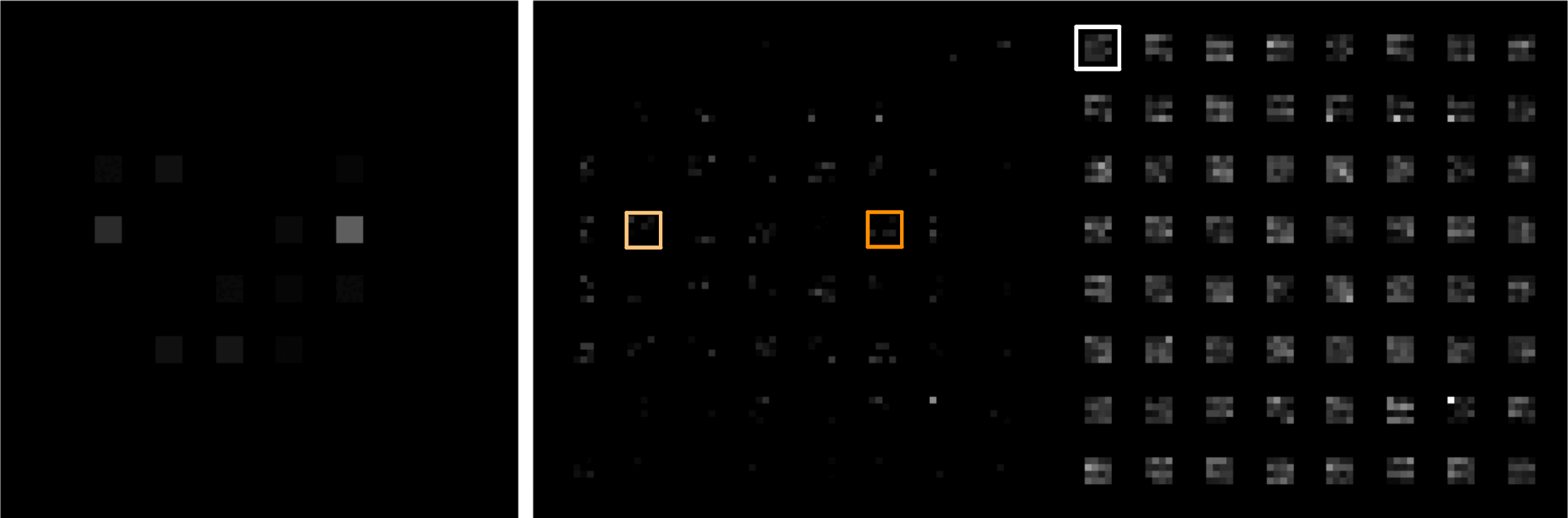}} \\ 
\multicolumn{3}{c}{\includegraphics[height=0.115\textheight]{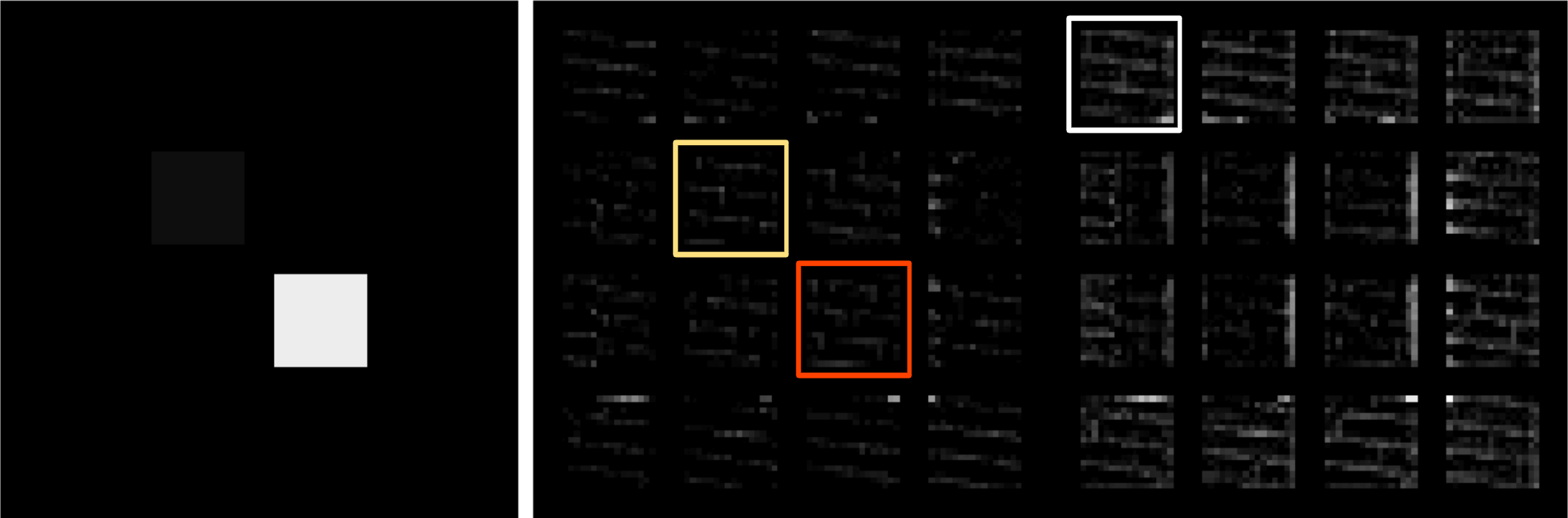}} \\ 
\multicolumn{3}{c}{\includegraphics[height=0.115\textheight]{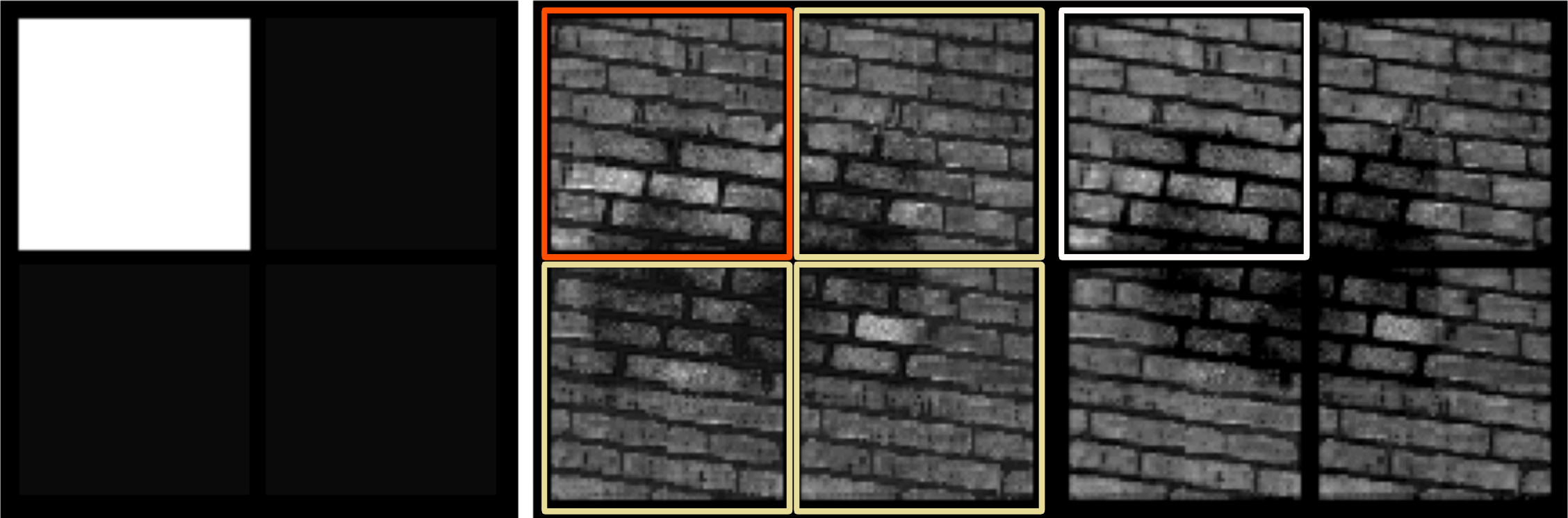}} 
\end{tabu}
\vspace{0.2cm}
\caption{Input/output visualization for all 5 Transformer blocks of the model (the images represent averaged feature maps along channel dimension, and black partitioning lines are added to distinguish the patches). At each stage, the attention map denotes the weights for computing the top-leftmost patch of the output (white square highlight) based on the input patches. More red highlighted squares on the inputs represent a higher impact on the output.}
\label{fig:vis2}
\vspace{-0.1cm}
\end{figure}

Our novel use of attention mechanisms enables new ways for patch mapping at varying scales and allows for interesting visualizations. We are able to visualize the attention to each considered patch at each stage of the proposed model. We can also visualize how the network progressively improves results with consecutive Transformer stages operating on varying scales. 
\cref{fig:vis2} presents the hierarchical patch mapping for one output patch. The figure illustrates how the proposed network progressively (top to bottom) refines the patch mapping process at different scales with coarse-to-fine and fine-back-to-coarse schemes.

\subsection{Ablation study}

We perform an ablation study to demonstrate the effectiveness of each component of our network. Our model is compared to: (1) a baseline architecture with three cascaded Transformer blocks; (2) the baseline (1) and our proposed progressive patches size reduction for later cascaded Transformers (also called pyramid architecture hereafter); (3) the hourglass backbone with two mirrored cascaded Transformer (2) and skip connections, without convolution downs and ups between Transformer blocks; (4) our proposed U-Attention net without the adversarial loss. 

We present the qualitative result of the ablation study in \cref{fig:ablation}. The baseline cannot handle multiple scales and patch sizes. The pyramid architecture (2) allows to attend to different patch sizes, but cannot unify the multi-scale structures properly. The simplified hourglass misses the convolutions between Transformer blocks and leads to inconsistent color shifts and worst structure alignment as seen in lines 2 and 4 of \cref{fig:ablation}.
Our use of GAN loss is particularly important to enhance the realism of the results of our network. Without the GAN loss, the borders of the image are well structured but suffer from a color and contrast shift. 

We provide the quantitative evaluation with L-PIPS and c-L-PIPS in \cref{tab:ablation} confirming the visual qualitative differences seen in \cref{fig:ablation}. More details of the ablated designs and additional synthesis results on a wide range of textures are provided in the supplemental materials.

\subsection{Limitations and future work} 

\cref{fig:lim} illustrates the limitations of our approach. As our method relies on patch feature matching, we cannot handle perspective as shown in the first row nor non-(semi)-periodic textures such as the unseen shapes of the bark in row 2 and the leather in row 3. Additionally, as we rely on an attention grid, our network may fail to align structures with limited training samples, creating small discontinuities visible in row 4. These discontinuities are however mitigated by our multi-scale approach and processing in feature space.

In some particularly adverse cases where the texture is not self-repeating enough and the coarse grid does not allow for relevant patches retrieval, some alignment artifacts can appear such as in the last row. In future work, it would be interesting to augment the pool of patches to attend to or increase the number of scales/stages of the hierarchical hourglass network for the attention mechanism to select the best alignment.

\section{Conclusions}
We propose a novel U-Attention network with multi-stage hierarchical hourglass vision Transformers enabling 2$\times$ texture synthesis on a broad range of texture patterns with a single trained network. Our approach uses features at multiple scales to synthesize patches with varying details while preserving the input texture appearance and runs in a few milliseconds. 
Our proposed network is a first step in leveraging the attention mechanism for texture synthesis, and we believe that the different components we demonstrate will be useful for broader research on attention-based image-to-image mapping.

\section*{Acknowledgment} 

We thank Jeff Fessler, Tamy Boubekeur, Paul Parneix, and Jerome Derel for insightful discussions.

{\small
\bibliographystyle{ieee_fullname}
\bibliography{main}
}

\newpage
\onecolumn

\section*{\Large Supplemental Material \\[0.5em] U-Attention to Textures: \\ Hierarchical Hourglass Vision Transformer for Universal Texture Synthesis}

\vspace{1em}
\setcounter{section}{0}
\def\thesection{\Alph{section}}
\setcounter{figure}{0}
\setcounter{table}{0}

This supplemental material presents: 
\begin{itemize}\itemsep=2pt
\item Details of our proposed network architecture (\cref{sec:supp_net})
\item Additional synthesis results on a wide range of textures (\cref{sec:supp_more}) 
\item Details of ablated designs and additional results for the ablation study (\cref{sec:supp_ab})
\item Texture synthesis of 128 $\rightarrow$ 256
(\cref{sec:supp_256})
\end{itemize}

\section{Network Architecture Details}
\label{sec:supp_net}

\Cref{tab:net} presents details of our proposed U-Attention network. We zero-pad the small texture exemplar (to the output size) as the input and solve the texture inpainting problem. Note that although \cref{tab:net} uses an overall input/output dimension of 128$\times$128 as an example, our network is not constrained to a particular size of input/output. Details of the temporal-patch adversarial loss \cite{chang2019free,zeng2020learning} we adopted are in \cref{tab:gloss}. 

\renewcommand{\arraystretch}{1.15}
\begin{table}[ht]
    \centering
    \scalebox{0.87}{
    \begin{tabular}{c|c|c|c|c|c|c|c}
         & Operation & Filter Size & \# Channels & Stride/Up Factor & Nonlinearity & Input & Output\\
        \hline
        \hline
         \multirow{2}{*}{Encoder} & 2D Conv & 3$\times$3 & 3 $\rightarrow$ 16 & 1 & LeakyReLU (0.2) & 128$\times$128 & 128$\times$128 \\
         & 2D Conv & 1$\times$1 & 16 $\rightarrow$ 16 & 1 & LeakyReLU (0.2) & 128$\times$128 & 128$\times$128 \\
         \hline
       \multirow{3}{*}{T-Block1} & Partition\_1 & - & - & - & - & 128$\times$128 & 4$\times$64$\times$64 \\
        & TransformerLayers\_1 & 1$\times$1, 3$\times$3 & 16 & 1 & LeakyReLU (0.2) & 4$\times$64$\times$64 & 4$\times$64$\times$64 \\
        & ArrangeBack\_1 & - & - & - & - & 4$\times$64$\times$64 & 128$\times$128 \\
        \hline
       \multirow{2}{*}{ConvDown1} & 2D Conv & 4$\times$4 & 16 $\rightarrow$ 64 & 2 & LeakyReLU (0.2) & 128$\times$128 & 64$\times$64 \\
        & 2D Conv & 1$\times$1 & 64 $\rightarrow$ 64 & 1 & LeakyReLU (0.2) & 64$\times$64 & 64$\times$64 \\
        \hline
        \multirow{3}{*}{T-Block2} & Partition\_2 & - & - & - & - & 64$\times$64 & 16$\times$16$\times$16 \\
        & TransformerLayers\_2 & 1$\times$1, 3$\times$3 & 64 & 1 & LeakyReLU (0.2) & 16$\times$16$\times$16 & 16$\times$16$\times$16 \\
        & ArrangeBack\_2 & - & - & - & - & 16$\times$16$\times$16 & 64$\times$64 \\
        \hline
        \multirow{2}{*}{ConvDown2} & 2D Conv & 4$\times$4 & 64 $\rightarrow$ 256 & 2 & LeakyReLU (0.2) & 64$\times$64 & 32$\times$32 \\
        & 2D Conv & 1$\times$1 & 256 $\rightarrow$ 256 & 1 & LeakyReLU (0.2) & 32$\times$32 & 32$\times$32\\
        \hline
        \multirow{3}{*}{T-Block3} & Partition\_3 & - & - & - & - & 32$\times$32 & 64$\times$4$\times$4 \\
        & TransformerLayers\_3 & 1$\times$1,  1$\times$1 & 256 & 1 & LeakyReLU (0.2) & 64$\times$4$\times$4 & 64$\times$4$\times$4 \\
        & ArrangeBack\_3 & - & - & - & - & 64$\times$4$\times$4 & 32$\times$32 \\
        \hline
       \multirow{3}{*}{ConvUp1} & BilinearUpSample & - & 256 & 2 & - & 32$\times$32 & 64$\times$64 \\
        & 2D Conv & 1$\times$1 & 256 $\rightarrow$ 64 & 1 & LeakyReLU (0.2) & 64$\times$64 & 64$\times$64 \\
        & 2D Conv & 1$\times$1 & 64 $\rightarrow$ 64 & 1 & LeakyReLU (0.2) & 64$\times$64 & 64$\times$64 \\
        \hline
       \multirow{2}{*}{ConvFuse1} & 2D Conv & 1$\times$1 & 128 $\rightarrow$ 64 & 1 & LeakyReLU (0.2) & 64$\times$64 & 64$\times$64 \\
        & 2D Conv & 1$\times$1 & 64 $\rightarrow$ 64 & 1 & LeakyReLU (0.2) & 64$\times$64 & 64$\times$64 \\
        \hline
       \multirow{3}{*}{T-Block4} & Partition\_2 & - & - & - & - & 64$\times$64 & 16$\times$16$\times$16 \\
        & TransformerLayers\_4 & 1$\times$1, 3$\times$3 & 64 & 1 & LeakyReLU (0.2) & 16$\times$16$\times$16 & 16$\times$16$\times$16 \\
        & ArrangeBack\_2 & - & - & - & - & 16$\times$16$\times$16 & 64$\times$64 \\
        \hline
       \multirow{3}{*}{ConvUp2} & BilinearUpSample & - & 64 & 2 & - & 64$\times$64 & 128$\times$128 \\
        & 2D Conv & 1$\times$1 & 64 $\rightarrow$ 16 & 1 & LeakyReLU (0.2) & 128$\times$128 & 128$\times$128 \\
        & 2D Conv & 1$\times$1 & 16 $\rightarrow$ 16 & 1 & LeakyReLU (0.2) & 128$\times$128 & 128$\times$128 \\
        \hline
       \multirow{2}{*}{ConvFuse2} & 2D Conv & 1$\times$1 & 32 $\rightarrow$ 16 & 1 & LeakyReLU (0.2) & 128$\times$128 & 128$\times$128 \\
        & 2D Conv & 1$\times$1 & 16 $\rightarrow$ 16 & 1 & LeakyReLU (0.2) & 128$\times$128 & 128$\times$128 \\
        \hline
       \multirow{3}{*}{T-Block5} & Partition\_1 & - & - & - & - & 128$\times$128 & 4$\times$64$\times$64 \\
        & TransformerLayers\_5 & 1$\times$1,  3$\times$3 & 16 & 1 & LeakyReLU (0.2) & 4$\times$64$\times$64 & 4$\times$64$\times$64 \\
        & ArrangeBack\_1 & - & - & - & - & 4$\times$64$\times$64 & 128$\times$128 \\
        \hline
       \multirow{2}{*}{Decoder}  & 2D Conv & 3$\times$3 & 16 $\rightarrow$ 3 & 1 & LeakyReLU (0.2) & 128$\times$128 & 128$\times$128 \\
         & 2D Conv & 1$\times$1 & 3 $\rightarrow$ 3 & 1 & Tanh & 128$\times$128 & 128$\times$128 \\
         \hline
    \end{tabular}
    }
    \vspace{1em}
    \caption{Details of the proposed U-Attention network architecture and parameters. The ``Input" and ``Output" denotes the spatial dimensions of the input and output at different stages of the network in forms of $H_i \times W_i$ for whole images or $P_i^2 \times \frac{H_i}{P_i} \times \frac{W_i}{P_i}$ for a sequence of $P_i^2$ patches.  ``ConvFuse1" fuses outputs of ``T-Block2" and ``ConvUp1" to be the input of ``T-Block4". Similarly, ``ConvFuse2" fuses outputs of ``T-Block1" and ``ConvUp2" to be the input of ``T-Block5". ``Partition\_$i$" represents the multi-scale partition for the proposed multi-scale attention in hierarchical hourglass Transformers. ``ArrangeBack\_$i$" denotes the operation that rearranges a sequence of patches back to whole feature maps. ``TransformerLayers\_$i$'' are composed of two stacked Transformers.
    }
    \label{tab:net}
\end{table}

\begin{table}
\centering   
\scalebox{0.89}{
\begin{tabular}{c|c|c|c|c} 
    Operation & Filter Size &\# Channels & Stride & Nonlinearity \\
    \hline\hline
    3D Conv & 3$\times$3$\times$3 & 3$\rightarrow$32 & (1, 1, 1) & LeakyReLU (0.2) \\ 
    3D Conv & 3$\times$3$\times$3 & 32$\rightarrow$64 & (1, 1, 1) & LeakyReLU (0.2) \\ 
    3D Conv & 3$\times$3$\times$3 & 64$\rightarrow$128 & (1, 1, 1) & LeakyReLU (0.2) \\ 
    3D Conv & 3$\times$3$\times$3 & 128$\rightarrow$128 & (1, 1, 1) & LeakyReLU (0.2) \\ 
    3D Conv & 3$\times$3$\times$3 & 128$\rightarrow$128 & (1, 1, 1) & LeakyReLU (0.2) \\ 
    3D Conv & 3$\times$3$\times$3 & 128$\rightarrow$128 & (1, 1, 1) & - \\ 
    \hline
    \end{tabular} 
    }
    \vspace{1em}
    \caption{Details of the temporal-patch adversarial loss (T-PatchGAN) \cite{chang2019free,zeng2020learning}. The 3D convolution layers for T-PatchGAN use spectral normalization to stabilize training \cite{chang2019free,zeng2020learning}.}
    \vspace{-1em}
    \label{tab:gloss}
    \end{table}   

\renewcommand{\arraystretch}{1}
\section{Additional Synthesis Results}
\label{sec:supp_more}

This section presents additional synthesis results of the proposed model compared to other methods and input-output pairs of a wide range of textures using the proposed model.

\subsection{Comparisons to other methods}

\begin{figure*}[ht]
\centering
\scalebox{0.72}{
\addtolength{\tabcolsep}{-4pt}
\begin{tabu}{cccccccc}
\rowfont{\Large} 
Input & Ours & WCT & pix2pixHD & Self Tuning & Texture CNN$^{*}$ & Naive Tiling  & Ground Truth 
\\ 
\includegraphics[width=0.09\textwidth]{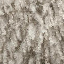} &
\includegraphics[width=0.18\textwidth]{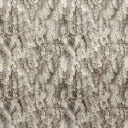} &
\includegraphics[width=0.18\textwidth]{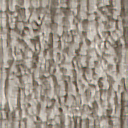} &
\includegraphics[width=0.18\textwidth]{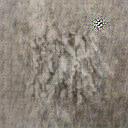} & 
\includegraphics[width=0.18\textwidth]{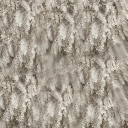} &
\includegraphics[width=0.18\textwidth]{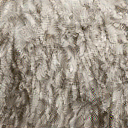} &
\includegraphics[width=0.18\textwidth]{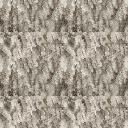} &
\includegraphics[width=0.18\textwidth]{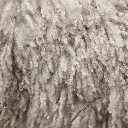} \\
\includegraphics[width=0.09\textwidth]{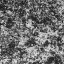} &
\includegraphics[width=0.18\textwidth]{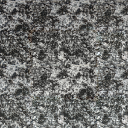} &
\includegraphics[width=0.18\textwidth]{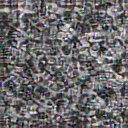} &
\includegraphics[width=0.18\textwidth]{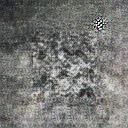} & 
\includegraphics[width=0.18\textwidth]{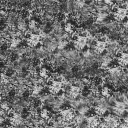} &
\includegraphics[width=0.18\textwidth]{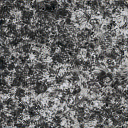} &
\includegraphics[width=0.18\textwidth]{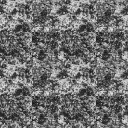} &
\includegraphics[width=0.18\textwidth]{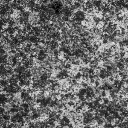} \\
\includegraphics[width=0.09\textwidth]{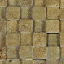} &
\includegraphics[width=0.18\textwidth]{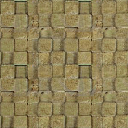} &
\includegraphics[width=0.18\textwidth]{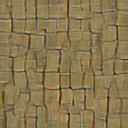} &
\includegraphics[width=0.18\textwidth]{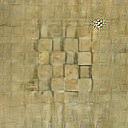} & 
\includegraphics[width=0.18\textwidth]{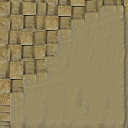} &
\includegraphics[width=0.18\textwidth]{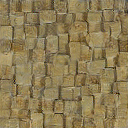} &
\includegraphics[width=0.18\textwidth]{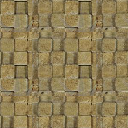} &
\includegraphics[width=0.18\textwidth]{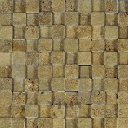} \\
\includegraphics[width=0.09\textwidth]{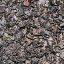} &
\includegraphics[width=0.18\textwidth]{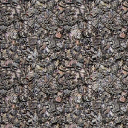} &
\includegraphics[width=0.18\textwidth]{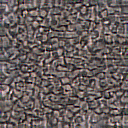} &
\includegraphics[width=0.18\textwidth]{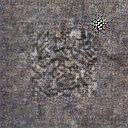} & 
\includegraphics[width=0.18\textwidth]{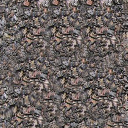} &
\includegraphics[width=0.18\textwidth]{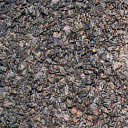} &
\includegraphics[width=0.18\textwidth]{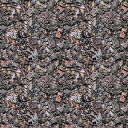} &
\includegraphics[width=0.18\textwidth]{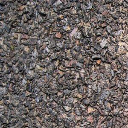} \\
\includegraphics[width=0.09\textwidth]{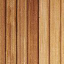} &
\includegraphics[width=0.18\textwidth]{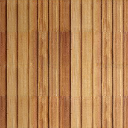} &
\includegraphics[width=0.18\textwidth]{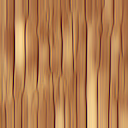} &
\includegraphics[width=0.18\textwidth]{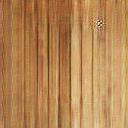} & 
\includegraphics[width=0.18\textwidth]{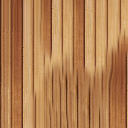} &
\includegraphics[width=0.18\textwidth]{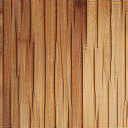} &
\includegraphics[width=0.18\textwidth]{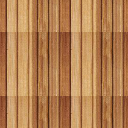} &
\includegraphics[width=0.18\textwidth]{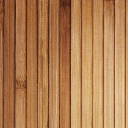} \\
\includegraphics[width=0.09\textwidth]{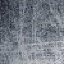} &
\includegraphics[width=0.18\textwidth]{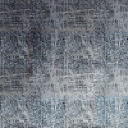} &
\includegraphics[width=0.18\textwidth]{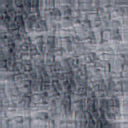} &
\includegraphics[width=0.18\textwidth]{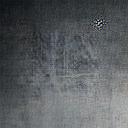} & 
\includegraphics[width=0.18\textwidth]{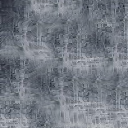} &
\includegraphics[width=0.18\textwidth]{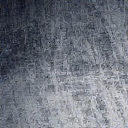} &
\includegraphics[width=0.18\textwidth]{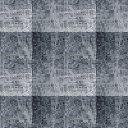} &
\includegraphics[width=0.18\textwidth]{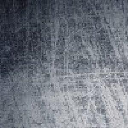} \\
\includegraphics[width=0.09\textwidth]{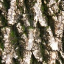} &
\includegraphics[width=0.18\textwidth]{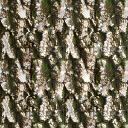} &
\includegraphics[width=0.18\textwidth]{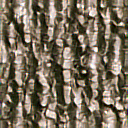} &
\includegraphics[width=0.18\textwidth]{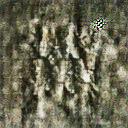} & 
\includegraphics[width=0.18\textwidth]{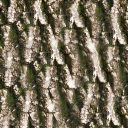} &
\includegraphics[width=0.18\textwidth]{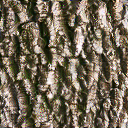} &
\includegraphics[width=0.18\textwidth]{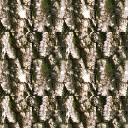} &
\includegraphics[width=0.18\textwidth]{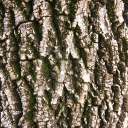} \\
\includegraphics[width=0.09\textwidth]{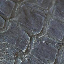} &
\includegraphics[width=0.18\textwidth]{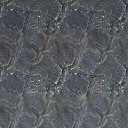} &
\includegraphics[width=0.18\textwidth]{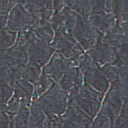} &
\includegraphics[width=0.18\textwidth]{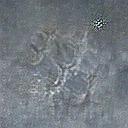} & 
\includegraphics[width=0.18\textwidth]{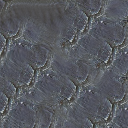} &
\includegraphics[width=0.18\textwidth]{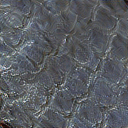} &
\includegraphics[width=0.18\textwidth]{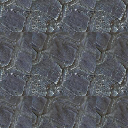} &
\includegraphics[width=0.18\textwidth]{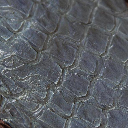} \\
\includegraphics[width=0.09\textwidth]{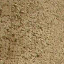} &
\includegraphics[width=0.18\textwidth]{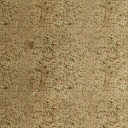} &
\includegraphics[width=0.18\textwidth]{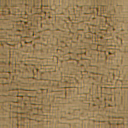} &
\includegraphics[width=0.18\textwidth]{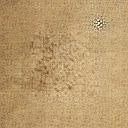} & 
\includegraphics[width=0.18\textwidth]{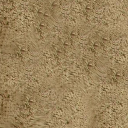} &
\includegraphics[width=0.18\textwidth]{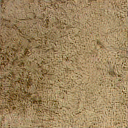} &
\includegraphics[width=0.18\textwidth]{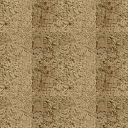} &
\includegraphics[width=0.18\textwidth]{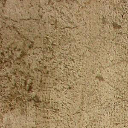}
\end{tabu}
}
\caption{Comparison to previous works demonstrates that our approach can handle textures of varying patterns and structures, and preserve the coherent color shift and structural details.
}
\label{fig:more_comp}
\vspace{-0.5cm}
\end{figure*}

\Cref{fig:more_comp} provides additional comparison results of the proposed approach to previous works.

\subsection{Results of our model for different textures}

\begin{figure*}[ht]
\centering
\scalebox{0.72}{
\addtolength{\tabcolsep}{-4pt}
\begin{tabu}{cccccccccc}
\rowfont{\Large} 
Input & Ours & Input & Ours & Input & Ours & Input & Ours & Input & Ours \\ 
\includegraphics[width=0.09\textwidth]{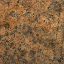} &
\includegraphics[width=0.18\textwidth]{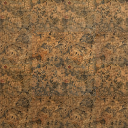} &
\includegraphics[width=0.09\textwidth]{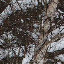} &
\includegraphics[width=0.18\textwidth]{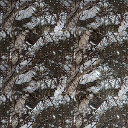} & 
\includegraphics[width=0.09\textwidth]{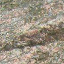} &
\includegraphics[width=0.18\textwidth]{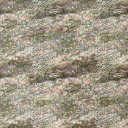} &
\includegraphics[width=0.09\textwidth]{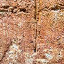} &
\includegraphics[width=0.18\textwidth]{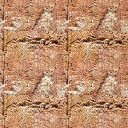} &
\includegraphics[width=0.09\textwidth]{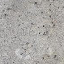} &
\includegraphics[width=0.18\textwidth]{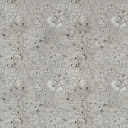} 
\\
\includegraphics[width=0.09\textwidth]{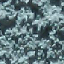} &
\includegraphics[width=0.18\textwidth]{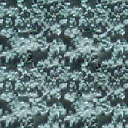} &
\includegraphics[width=0.09\textwidth]{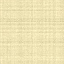} &
\includegraphics[width=0.18\textwidth]{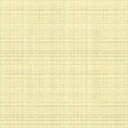} & 
\includegraphics[width=0.09\textwidth]{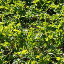} &
\includegraphics[width=0.18\textwidth]{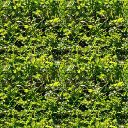} &
\includegraphics[width=0.09\textwidth]{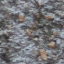} &
\includegraphics[width=0.18\textwidth]{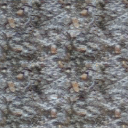} &
\includegraphics[width=0.09\textwidth]{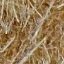} &
\includegraphics[width=0.18\textwidth]{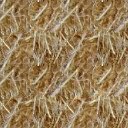} 
\\
\includegraphics[width=0.09\textwidth]{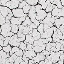} &
\includegraphics[width=0.18\textwidth]{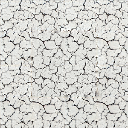} &
\includegraphics[width=0.09\textwidth]{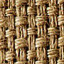} &
\includegraphics[width=0.18\textwidth]{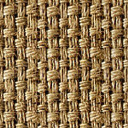} & 
\includegraphics[width=0.09\textwidth]{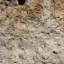} &
\includegraphics[width=0.18\textwidth]{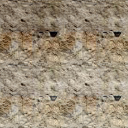} &
\includegraphics[width=0.09\textwidth]{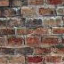} &
\includegraphics[width=0.18\textwidth]{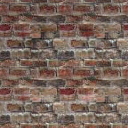} &
\includegraphics[width=0.09\textwidth]{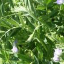} &
\includegraphics[width=0.18\textwidth]{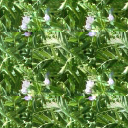} 
\\
\includegraphics[width=0.09\textwidth]{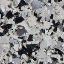} &
\includegraphics[width=0.18\textwidth]{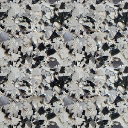} &
\includegraphics[width=0.09\textwidth]{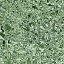} &
\includegraphics[width=0.18\textwidth]{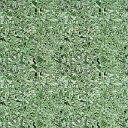} & 
\includegraphics[width=0.09\textwidth]{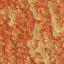} &
\includegraphics[width=0.18\textwidth]{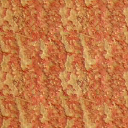} &
\includegraphics[width=0.09\textwidth]{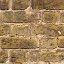} &
\includegraphics[width=0.18\textwidth]{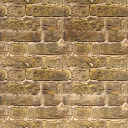} &
\includegraphics[width=0.09\textwidth]{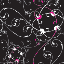} &
\includegraphics[width=0.18\textwidth]{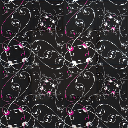} 
\\
\includegraphics[width=0.09\textwidth]{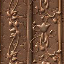} &
\includegraphics[width=0.18\textwidth]{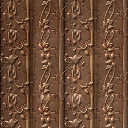} &
\includegraphics[width=0.09\textwidth]{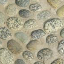} &
\includegraphics[width=0.18\textwidth]{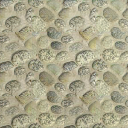} & 
\includegraphics[width=0.09\textwidth]{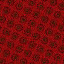} &
\includegraphics[width=0.18\textwidth]{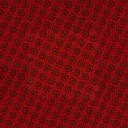} &
\includegraphics[width=0.09\textwidth]{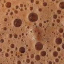} &
\includegraphics[width=0.18\textwidth]{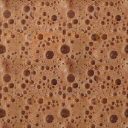} &
\includegraphics[width=0.09\textwidth]{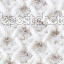} &
\includegraphics[width=0.18\textwidth]{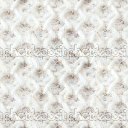} 
\\
\includegraphics[width=0.09\textwidth]{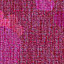} &
\includegraphics[width=0.18\textwidth]{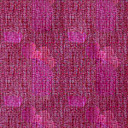} &
\includegraphics[width=0.09\textwidth]{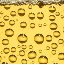} &
\includegraphics[width=0.18\textwidth]{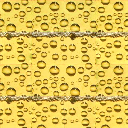} & 
\includegraphics[width=0.09\textwidth]{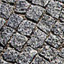} &
\includegraphics[width=0.18\textwidth]{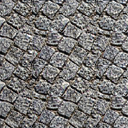} &
\includegraphics[width=0.09\textwidth]{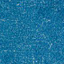} &
\includegraphics[width=0.18\textwidth]{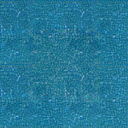} &
\includegraphics[width=0.09\textwidth]{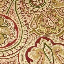} &
\includegraphics[width=0.18\textwidth]{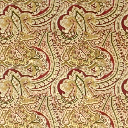} 
\\
\includegraphics[width=0.09\textwidth]{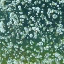} &
\includegraphics[width=0.18\textwidth]{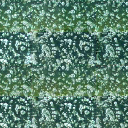} &
\includegraphics[width=0.09\textwidth]{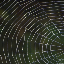} &
\includegraphics[width=0.18\textwidth]{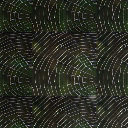} & 
\includegraphics[width=0.09\textwidth]{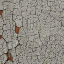} &
\includegraphics[width=0.18\textwidth]{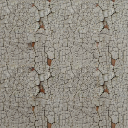} &
\includegraphics[width=0.09\textwidth]{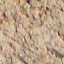} &
\includegraphics[width=0.18\textwidth]{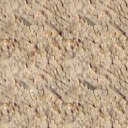} &
\includegraphics[width=0.09\textwidth]{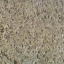} &
\includegraphics[width=0.18\textwidth]{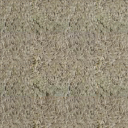} 
\\
\includegraphics[width=0.09\textwidth]{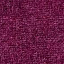} &
\includegraphics[width=0.18\textwidth]{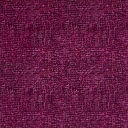} &
\includegraphics[width=0.09\textwidth]{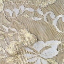} &
\includegraphics[width=0.18\textwidth]{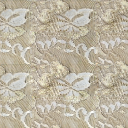} & 
\includegraphics[width=0.09\textwidth]{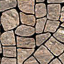} &
\includegraphics[width=0.18\textwidth]{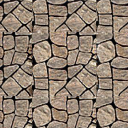} &
\includegraphics[width=0.09\textwidth]{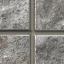} &
\includegraphics[width=0.18\textwidth]{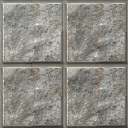} &
\includegraphics[width=0.09\textwidth]{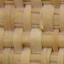} &
\includegraphics[width=0.18\textwidth]{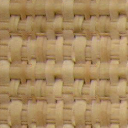} 
\\
\includegraphics[width=0.09\textwidth]{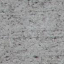} &
\includegraphics[width=0.18\textwidth]{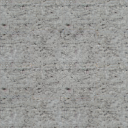} &
\includegraphics[width=0.09\textwidth]{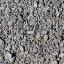} &
\includegraphics[width=0.18\textwidth]{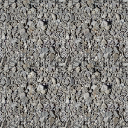} & 
\includegraphics[width=0.09\textwidth]{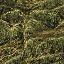} &
\includegraphics[width=0.18\textwidth]{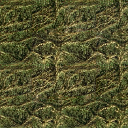} &
\includegraphics[width=0.09\textwidth]{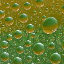} &
\includegraphics[width=0.18\textwidth]{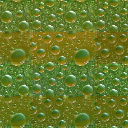} &
\includegraphics[width=0.09\textwidth]{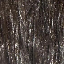} &
\includegraphics[width=0.18\textwidth]{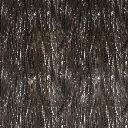}
\end{tabu}
}
\caption{Synthesis results of our U-Attention network demonstrate that our approach generalizes to a broad range of textures with varieties of randomness and structure using one trained network.
}
\label{fig:more_results1}
\vspace{-0.5cm}
\end{figure*}

\begin{figure*}[htbp]
\centering
\scalebox{0.72}{
\addtolength{\tabcolsep}{-4pt}
\begin{tabu}{cccccccccc}
\rowfont{\Large} 
Input & Ours & Input & Ours & Input & Ours & Input & Ours & Input & Ours \\
\includegraphics[width=0.09\textwidth]{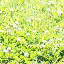} &
\includegraphics[width=0.18\textwidth]{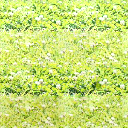} &
\includegraphics[width=0.09\textwidth]{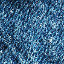} &
\includegraphics[width=0.18\textwidth]{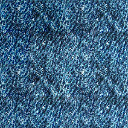} & 
\includegraphics[width=0.09\textwidth]{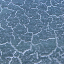} &
\includegraphics[width=0.18\textwidth]{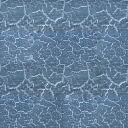} &
\includegraphics[width=0.09\textwidth]{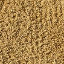} &
\includegraphics[width=0.18\textwidth]{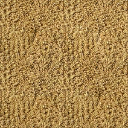} &
\includegraphics[width=0.09\textwidth]{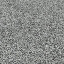} &
\includegraphics[width=0.18\textwidth]{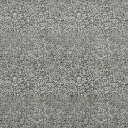} 
\\
\includegraphics[width=0.09\textwidth]{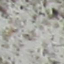} &
\includegraphics[width=0.18\textwidth]{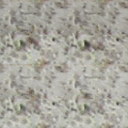} &
\includegraphics[width=0.09\textwidth]{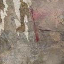} &
\includegraphics[width=0.18\textwidth]{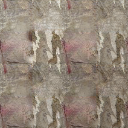} & 
\includegraphics[width=0.09\textwidth]{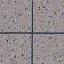} &
\includegraphics[width=0.18\textwidth]{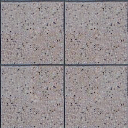} &
\includegraphics[width=0.09\textwidth]{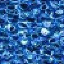} &
\includegraphics[width=0.18\textwidth]{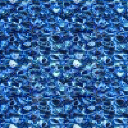} &
\includegraphics[width=0.09\textwidth]{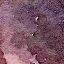} &
\includegraphics[width=0.18\textwidth]{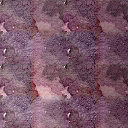} 
\\
\includegraphics[width=0.09\textwidth]{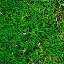} &
\includegraphics[width=0.18\textwidth]{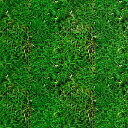} &
\includegraphics[width=0.09\textwidth]{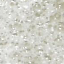} &
\includegraphics[width=0.18\textwidth]{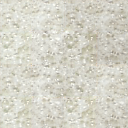} & 
\includegraphics[width=0.09\textwidth]{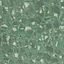} &
\includegraphics[width=0.18\textwidth]{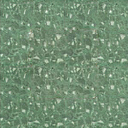} &
\includegraphics[width=0.09\textwidth]{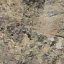} &
\includegraphics[width=0.18\textwidth]{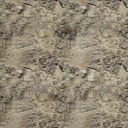} &
\includegraphics[width=0.09\textwidth]{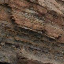} &
\includegraphics[width=0.18\textwidth]{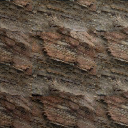} 
\\
\includegraphics[width=0.09\textwidth]{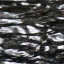} &
\includegraphics[width=0.18\textwidth]{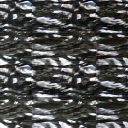} &
\includegraphics[width=0.09\textwidth]{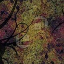} &
\includegraphics[width=0.18\textwidth]{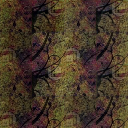} & 
\includegraphics[width=0.09\textwidth]{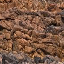} &
\includegraphics[width=0.18\textwidth]{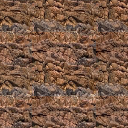} &
\includegraphics[width=0.09\textwidth]{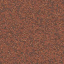} &
\includegraphics[width=0.18\textwidth]{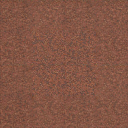} &
\includegraphics[width=0.09\textwidth]{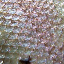} &
\includegraphics[width=0.18\textwidth]{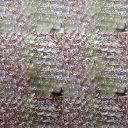} 
\\
\includegraphics[width=0.09\textwidth]{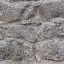} &
\includegraphics[width=0.18\textwidth]{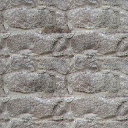} &
\includegraphics[width=0.09\textwidth]{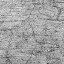} &
\includegraphics[width=0.18\textwidth]{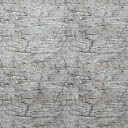} & 
\includegraphics[width=0.09\textwidth]{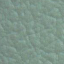} &
\includegraphics[width=0.18\textwidth]{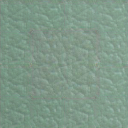} &
\includegraphics[width=0.09\textwidth]{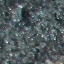} &
\includegraphics[width=0.18\textwidth]{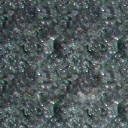} &
\includegraphics[width=0.09\textwidth]{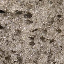} &
\includegraphics[width=0.18\textwidth]{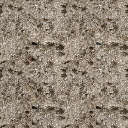} 
\\
\includegraphics[width=0.09\textwidth]{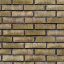} &
\includegraphics[width=0.18\textwidth]{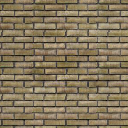} &
\includegraphics[width=0.09\textwidth]{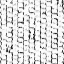} &
\includegraphics[width=0.18\textwidth]{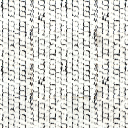} & 
\includegraphics[width=0.09\textwidth]{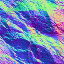} &
\includegraphics[width=0.18\textwidth]{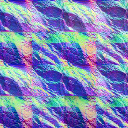} &
\includegraphics[width=0.09\textwidth]{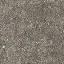} &
\includegraphics[width=0.18\textwidth]{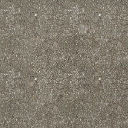} &
\includegraphics[width=0.09\textwidth]{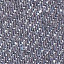} &
\includegraphics[width=0.18\textwidth]{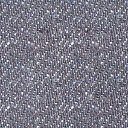} 
\\
\includegraphics[width=0.09\textwidth]{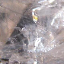} &
\includegraphics[width=0.18\textwidth]{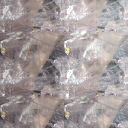} &
\includegraphics[width=0.09\textwidth]{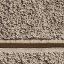} &
\includegraphics[width=0.18\textwidth]{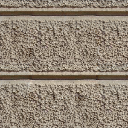} & 
\includegraphics[width=0.09\textwidth]{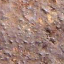} &
\includegraphics[width=0.18\textwidth]{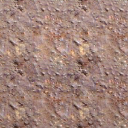} &
\includegraphics[width=0.09\textwidth]{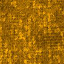} &
\includegraphics[width=0.18\textwidth]{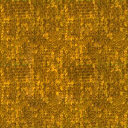} &
\includegraphics[width=0.09\textwidth]{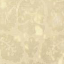} &
\includegraphics[width=0.18\textwidth]{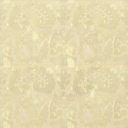} 
\\
\includegraphics[width=0.09\textwidth]{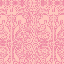} &
\includegraphics[width=0.18\textwidth]{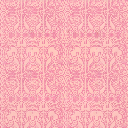} &
\includegraphics[width=0.09\textwidth]{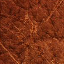} &
\includegraphics[width=0.18\textwidth]{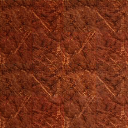} & 
\includegraphics[width=0.09\textwidth]{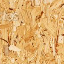} &
\includegraphics[width=0.18\textwidth]{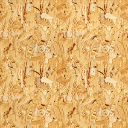} &
\includegraphics[width=0.09\textwidth]{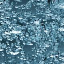} &
\includegraphics[width=0.18\textwidth]{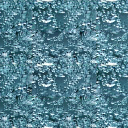} &
\includegraphics[width=0.09\textwidth]{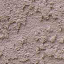} &
\includegraphics[width=0.18\textwidth]{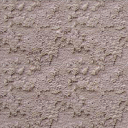} 
\\
\includegraphics[width=0.09\textwidth]{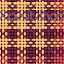} &
\includegraphics[width=0.18\textwidth]{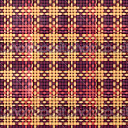} &
\includegraphics[width=0.09\textwidth]{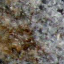} &
\includegraphics[width=0.18\textwidth]{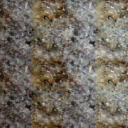} & 
\includegraphics[width=0.09\textwidth]{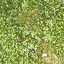} &
\includegraphics[width=0.18\textwidth]{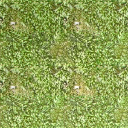} &
\includegraphics[width=0.09\textwidth]{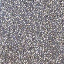} &
\includegraphics[width=0.18\textwidth]{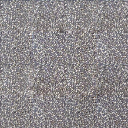} &
\includegraphics[width=0.09\textwidth]{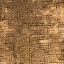} &
\includegraphics[width=0.18\textwidth]{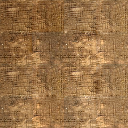}
\end{tabu}
}
\caption{More synthesis results of our U-Attention network demonstrate that our approach generalizes to a broad range of textures with varieties of randomness and structure using one trained network.
}
\label{fig:more_results2}
\vspace{-0.5cm}
\end{figure*}

\Cref{fig:more_results1} and \Cref{fig:more_results2} present more test results from our single trained network.

\section{Ablation Study}
\label{sec:supp_ab}

This section presents details about the ``pyramid" and ``simplified hourglass" network architectures introduced in the ablation study, along  with additional synthesis results.

\subsection{Pyramid and simplified hourglass architectures}

\begin{figure*}
\centering
     \includegraphics[width=0.9\textwidth]{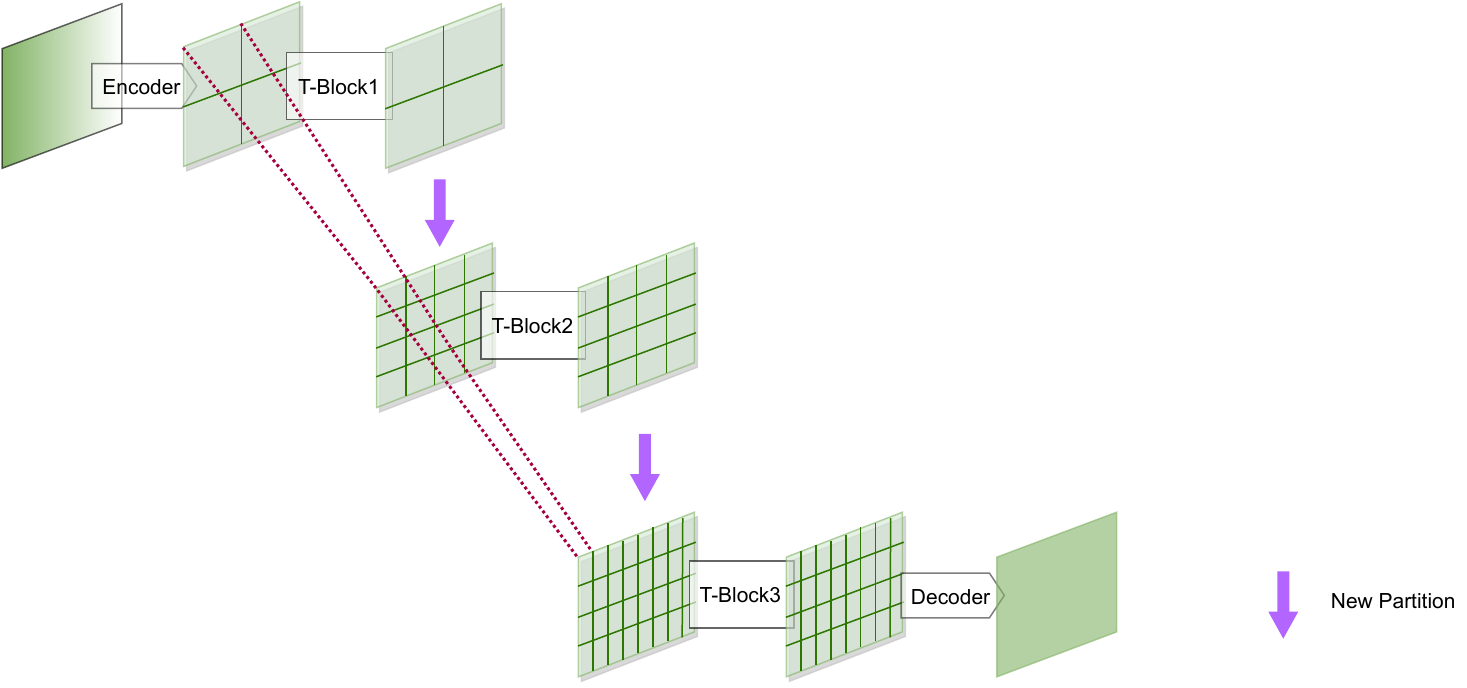}
   \caption{
   The ``pyramid" network.
   Each Transformer block takes the whole feature maps as inputs, and we partition the feature maps to be sequences of patches of shrinking (2$\times$ smaller in each dimension) spatial extents for consecutive stages of the network.
   }\label{fig:net_P}
   \vspace{-0.5cm}
\end{figure*}

We demonstrate the effective components of the proposed network for ablation study. The ``pyramid" network shown in \cref{fig:net_P} represents a multi-stage hierarchical Transformer network with an encoder, 3 Transformer blocks, and a decoder. The Transformer blocks take sequences of patches of feature maps as inputs.
We introduce a multi-scale partition of the feature map between Transformer blocks to form input patches of different scales for different Transformers.
Specifically, the input patch size presents a pyramid-like shrinking pattern with 2$\times$ smaller spatial extent between Transformer stages.

This ``pyramid'' network design establishes a coarse-to-fine patch mapping scheme that combines attention between patches of varying scales for global pattern outline and local refinements of detailed structures. In the extreme form of the pyramid, the last stage of the Transformer would take sequences of patches of size 1$\times$1 as input, and the results from the pyramid network might be improved at the expense of computation.

\begin{figure*}
\centering
     \includegraphics[width=0.9\textwidth]{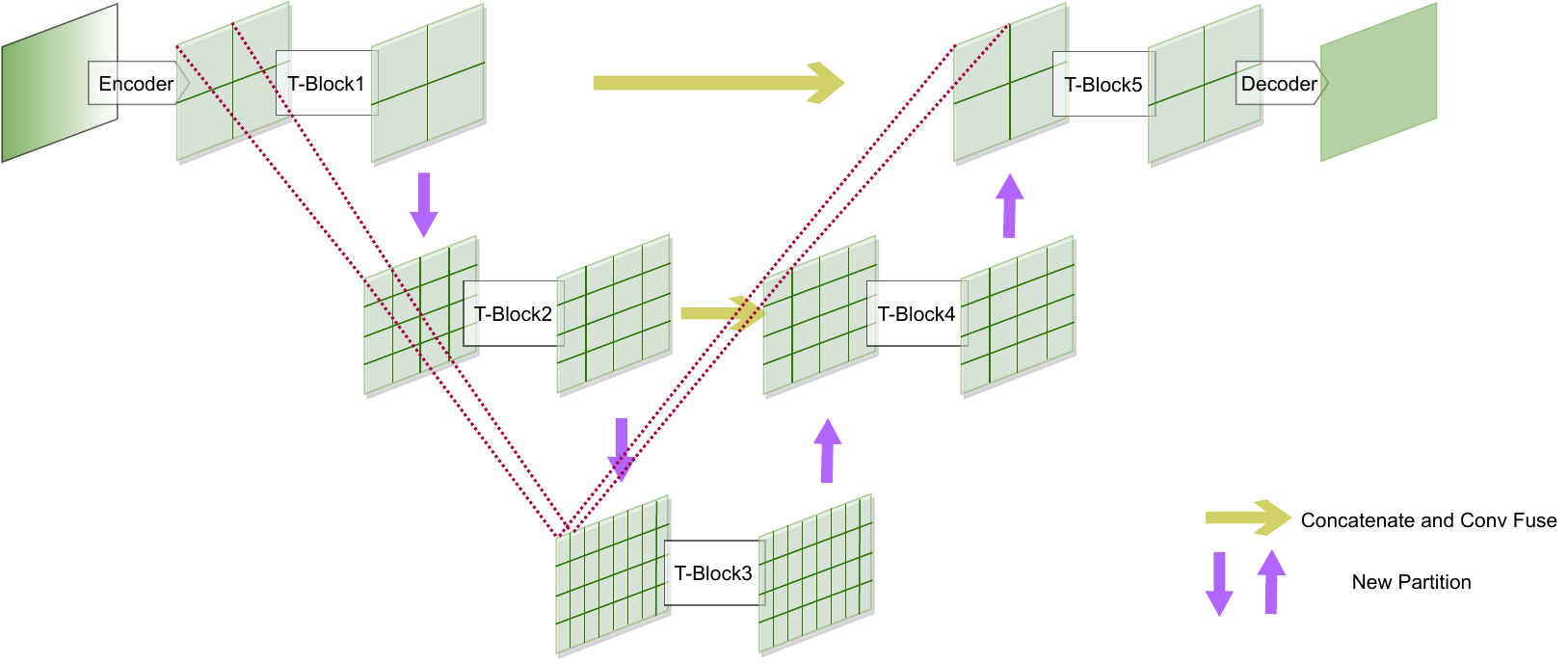}
   \caption{
   The ``simplified hourglass" network.
   Each Transformer block takes the whole feature maps as inputs, and we partition the feature maps to be sequences of patches of shrinking or enlarging spatial extents (2$\times$ smaller or larger in each dimension) for different stages of the hourglass network.
   }\label{fig:net_H}
   \vspace{-0.5cm}
\end{figure*}

The ``simplified hourglass" in \cref{fig:net_H}  concatenates the pyramid network with a mirrored pyramid network and is complemented with skip connections to propagate and fuse information. The mirrored pyramid network at the second half of the simplified hourglass takes progressively larger patches (2$\times$ larger in each spatial dimension between consecutive Transformer stages) with shorter sequence lengths as inputs for later Transformer stages. We also add skip connections that propagate output from an earlier stage to a later stage of the network, concatenate and fuse the outputs from the two stages with two layers of 1$\times$1 convolutions to form the input of the next Transformer.

This hourglass network design further regularizes the smallest output patches from the bottleneck of the hourglass and helps resolve blocking artifacts caused by patch-based operations of the pyramid network. The hourglass-like scale change of the patches enables coarse-to-fine and fine-back-to-coarse mapping. 

Note that for both pyramid and simplified hourglass networks, the input feature map  (before partitioning and arranging the patches as a sequence) for each Transformer stage are of the same spatial extent and same channel dimension, while in our U-Attention network the spatial extent and channel dimension of the input feature maps do vary with depth, as shown in \cref{tab:net}.

\subsection{More results of ablation study}

\begin{figure*}
\vspace{-0.3cm}
\centering
\scalebox{0.64}{
\addtolength{\tabcolsep}{-5pt}
\begin{tabu}{cccccc}
\multicolumn{1}{c}{} & \multicolumn{5}{c}{} \\
\rowfont{\Large} Input & Baseline & Pyramid & Simplified Hourglass & No GAN Loss & Ours Full \\ 
\includegraphics[width=0.12\textwidth]{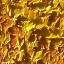} &
\includegraphics[width=0.24\textwidth]{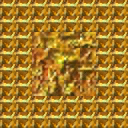} &
\includegraphics[width=0.24\textwidth]{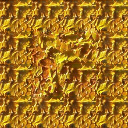} &
\includegraphics[width=0.24\textwidth]{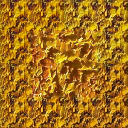} &
\includegraphics[width=0.24\textwidth]{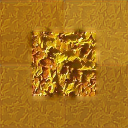} &
\includegraphics[width=0.24\textwidth]{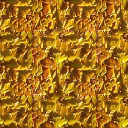} \\
\includegraphics[width=0.12\textwidth]{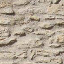} &
\includegraphics[width=0.24\textwidth]{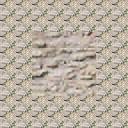} &
\includegraphics[width=0.24\textwidth]{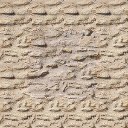} &
\includegraphics[width=0.24\textwidth]{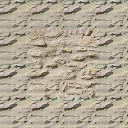} &
\includegraphics[width=0.24\textwidth]{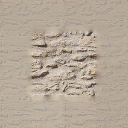} &
\includegraphics[width=0.24\textwidth]{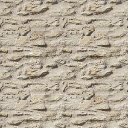} \\
\includegraphics[width=0.12\textwidth]{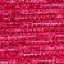} &
\includegraphics[width=0.24\textwidth]{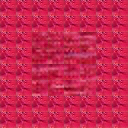} &
\includegraphics[width=0.24\textwidth]{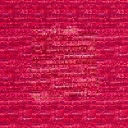} &
\includegraphics[width=0.24\textwidth]{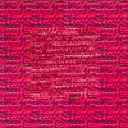} &
\includegraphics[width=0.24\textwidth]{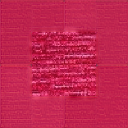} &
\includegraphics[width=0.24\textwidth]{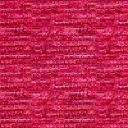} \\
\includegraphics[width=0.12\textwidth]{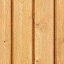} &
\includegraphics[width=0.24\textwidth]{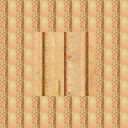} &
\includegraphics[width=0.24\textwidth]{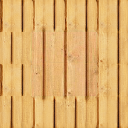} &
\includegraphics[width=0.24\textwidth]{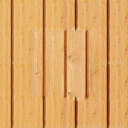} &
\includegraphics[width=0.24\textwidth]{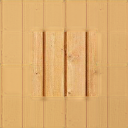} &
\includegraphics[width=0.24\textwidth]{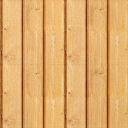} \\
\includegraphics[width=0.12\textwidth]{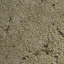} &
\includegraphics[width=0.24\textwidth]{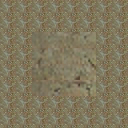} &
\includegraphics[width=0.24\textwidth]{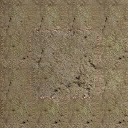} &
\includegraphics[width=0.24\textwidth]{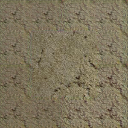} &
\includegraphics[width=0.24\textwidth]{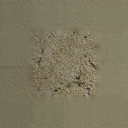} &
\includegraphics[width=0.24\textwidth]{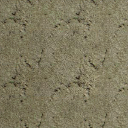} \\
\includegraphics[width=0.12\textwidth]{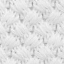} &
\includegraphics[width=0.24\textwidth]{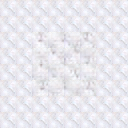} &
\includegraphics[width=0.24\textwidth]{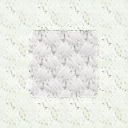} &
\includegraphics[width=0.24\textwidth]{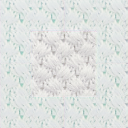} &
\includegraphics[width=0.24\textwidth]{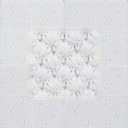} &
\includegraphics[width=0.24\textwidth]{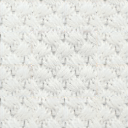} \\
\includegraphics[width=0.12\textwidth]{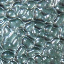} &
\includegraphics[width=0.24\textwidth]{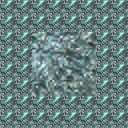} &
\includegraphics[width=0.24\textwidth]{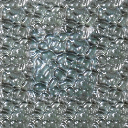} &
\includegraphics[width=0.24\textwidth]{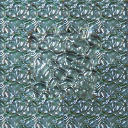} &
\includegraphics[width=0.24\textwidth]{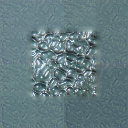} &
\includegraphics[width=0.24\textwidth]{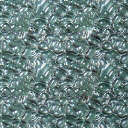} \\
\includegraphics[width=0.12\textwidth]{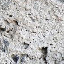} &
\includegraphics[width=0.24\textwidth]{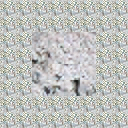} &
\includegraphics[width=0.24\textwidth]{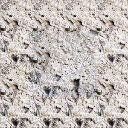} &
\includegraphics[width=0.24\textwidth]{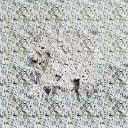} &
\includegraphics[width=0.24\textwidth]{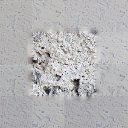} &
\includegraphics[width=0.24\textwidth]{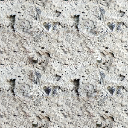} 
\end{tabu}
}
\caption{More qualitative synthesis results for our ablation study. 
}
\label{fig:more_ab}
\vspace{-0.2cm}
\end{figure*}

In \Cref{fig:more_ab} we show that our proposed network outperforms the multiple Transformer blocks baseline network. We further demonstrate the contribution of each component of our design through more examples to support our ablation study.

\section{128 to 256 Synthesis} 
\label{sec:supp_256}

\begin{figure*}
\centering
\scalebox{0.72}{
\addtolength{\tabcolsep}{-4pt}
\begin{tabu}{cccccccc}
\rowfont{\Large} 
Input & Ours & Input & Ours & Input & Ours \\ 
\includegraphics[width=0.15\textwidth]{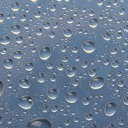} &
\includegraphics[width=0.3\textwidth]{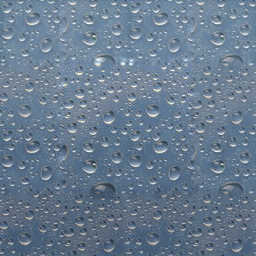} &
\includegraphics[width=0.15\textwidth]{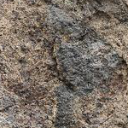} &
\includegraphics[width=0.3\textwidth]{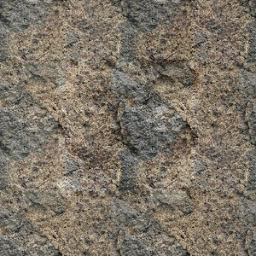} &
\includegraphics[width=0.15\textwidth]{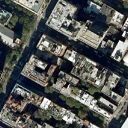} &
\includegraphics[width=0.3\textwidth]{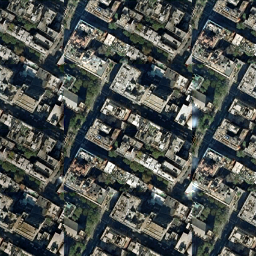} &
\\
\includegraphics[width=0.15\textwidth]{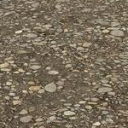} &
\includegraphics[width=0.3\textwidth]{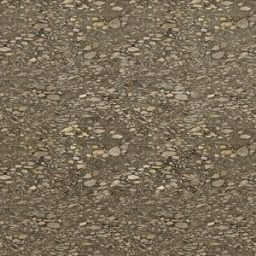} &
\includegraphics[width=0.15\textwidth]{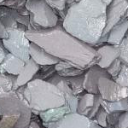} &
\includegraphics[width=0.3\textwidth]{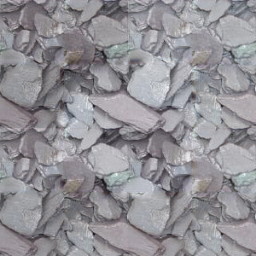} &
\includegraphics[width=0.15\textwidth]{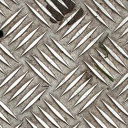} &
\includegraphics[width=0.3\textwidth]{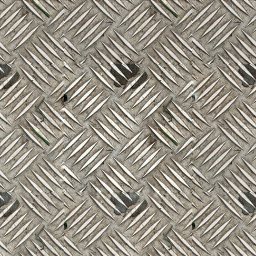} &
\\
\includegraphics[width=0.15\textwidth]{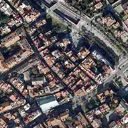} &
\includegraphics[width=0.3\textwidth]{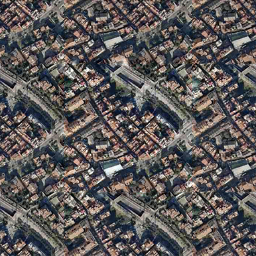} &
\includegraphics[width=0.15\textwidth]{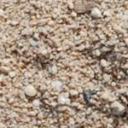} &
\includegraphics[width=0.3\textwidth]{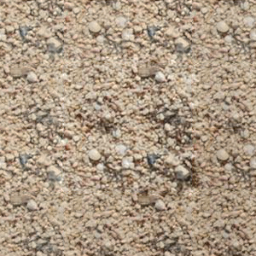} &
\includegraphics[width=0.15\textwidth]{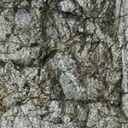} &
\includegraphics[width=0.3\textwidth]{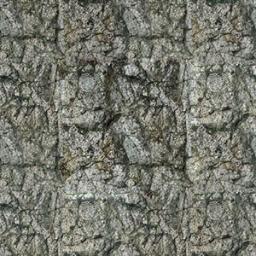} &
\end{tabu}
}
\caption{Higher-resolution synthesis results for image of size 128 to 256.
}
\label{fig:256_results}
\vspace{-0.5cm}
\end{figure*}

We fine-tune the network that is pre-trained for 64 to 128 synthesis with input and target images of larger size (256 instead of 128) for 40 epochs (about 4 days) and present 128 to 256 synthesis results. The 128 to 256 results in \cref{fig:256_results} are very similar to the quality of 64 to 128 with a higher resolution.

We believe that for larger texture synthesis, we could extend our proposed network to be composed of more stages of multi-scale attention networks and train the extended network for further improvement of performance.

\end{document}